\documentclass[letterpaper]{article} 
\usepackage[preprint]{aaai2027}  
\usepackage[hyphens]{url}  
\usepackage{graphicx} 
\usepackage{natbib}  
\usepackage{caption}
\usepackage{afterpage} 
\usepackage{algorithm}
\usepackage{algorithmic}

\usepackage{newfloat}
\usepackage{listings}
\DeclareCaptionStyle{ruled}{labelfont=normalfont,labelsep=colon,strut=off} 
\floatstyle{ruled}
\newfloat{listing}{tb}{lst}{}
\floatname{listing}{Listing}

\usepackage{booktabs}
\usepackage{makecell}
\usepackage{amssymb,amsmath}
\usepackage{array,tabularx}
\usepackage{xcolor}
\usepackage{colortbl}
\newcolumntype{Y}{>{\raggedright\arraybackslash}X}
\newcolumntype{C}{>{\centering\arraybackslash}X}
\definecolor{CVTBlue}{HTML}{2267B5}
\definecolor{CVTGreen}{HTML}{388A51}
\definecolor{CVTPurple}{HTML}{6544A5}
\definecolor{CVTRed}{HTML}{B72632}
\definecolor{CVTPaleBlue}{HTML}{EFF5FC}
\definecolor{CVTPaleGreen}{HTML}{EFF8F1}
\definecolor{CVTPaleOrange}{HTML}{FFF6EA}
\definecolor{CVTPalePurple}{HTML}{F5F1FC}
\definecolor{CVTInk}{HTML}{142033}
\lstdefinestyle{cvtprompt}{
  basicstyle=\scriptsize\ttfamily,
  breaklines=true,
  columns=fullflexible,
  showstringspaces=false,
  backgroundcolor=\color{CVTPaleBlue},
  frame=single,
  rulecolor=\color{CVTInk},
  framerule=.65pt,
  framesep=3.5pt,
  xleftmargin=0pt,
  xrightmargin=0pt,
  aboveskip=0pt,
  belowskip=3pt
}
\newcommand{\promptheader}[1]{%
  \par\smallskip\noindent
  \colorbox{CVTInk}{\parbox{\dimexpr\columnwidth-2\fboxsep\relax}{%
    \color{white}\bfseries\strut #1}}\par\nobreak}
\newenvironment{promptblock}[1]{%
  \par\smallskip\noindent\promptheader{#1}%
}{%
  \par\smallskip
}

\title{CVT-Bench: Probing Spatial-State Integrity through Counterfactual Viewpoint Transformations\thanks{Project page: \protect\url{https://shanmukha-here.github.io/CVT-Bench/}}}
\author{
Shanmukha Vellamcheti,
Uday Kiran Kothapalli,
Disharee Bhowmick,
Sathyanarayanan N. Aakur
}
\affiliations{
Department of Computer Science and Software Engineering, Auburn University\\
Auburn, AL 36849, USA\\
\{szv0080, uzk0006, dzb0110, san0028\}@auburn.edu
}

\begin{document}

\maketitle

\begin{abstract}
Multimodal large language models (MLLMs) perform strongly on isolated spatial tasks, but whether their predictions remain persistent and mutually coherent across viewpoints and competing scenes is unclear. We formalize this behavioral property as \emph{spatial-state integrity} and introduce CVT-Bench, a factorial diagnostic suite spanning two domains (CVT-Synthetic, CVT-Real), four context regimes (CVT-Isolated, irrelevant filler, attribute filler, CVT-Competing), three representations (Image, Text/BBox, Scene Graph), and ten viewpoint conditions (nine azimuths from \(0^\circ\) to \(360^\circ\) at \(45^\circ\) increments, plus Top) with target views withheld. Across 200 scenes and 11,833 relational queries, we measure counterfactual accuracy, cycle consistency, the normalized survival coefficient, spatial realizability, and context-specific interference. Five state-of-the-art MLLMs exhibit rapid persistence loss and frequently produce jointly unrealizable states despite high local accuracy, with failures amplified by competing contexts and natural-scene complexity. Matched irrelevant and attribute filler controls show that prompt length, position, or loss of scene access alone are insufficient explanations. On average, Text/BBox improves accuracy, persistence, and realizability, while Scene Graph provides complementary gains; neither eliminates the instability. Thus, isolated spatial accuracy substantially overestimates robustness, establishing spatial-state integrity as a distinct evaluation target. The complete benchmark and codebase will be publicly released.
\end{abstract}


\section{Introduction}
Humans can reason about a scene beyond the observed viewpoint by mentally adopting an alternative perspective and transforming observer-relative spatial relations. This ability, commonly studied as Level-2 visual perspective taking~\cite{michelon2006two}, requires the underlying scene to remain coherent even as relations such as left, right, front, and behind change with viewpoint. We define this behavioral property as \textbf{spatial-state integrity}, the ability to preserve and transform coherent spatial predictions across related viewpoints while keeping them distinct from competing scene contexts. This property is important for artificial agents that must reuse spatial knowledge inferred from a single observation as viewpoints and surrounding contexts change. Yet existing MLLM benchmarks typically evaluate only whether a model answers an isolated spatial question correctly. Such one-shot accuracy reflects local task competence, but not whether related predictions remain persistent and mutually compatible. We argue that correct, jointly realizable responses across counterfactual viewpoints provide a stronger behavioral signature of spatial-state integrity.

Assessing spatial-state integrity requires disentangling the underlying scene from the observer-relative frames used to describe it. Because a viewpoint change alters relational descriptions without changing the physical world, the same scene must be probed across controlled angular disparities under known geometric constraints. Interpreting failures is nevertheless difficult because performance may degrade through visual grounding errors, generic context effects, loss of access to scene information, or the accumulation of independent errors rather than instability in spatial reasoning itself. A rigorous evaluation must vary viewpoint and context independently, compare probing under CVT-Isolated and CVT-Competing, introduce matched controls for context length and scene access, and measure whether predictions remain accurate, persistent, and realizable across controlled and realistic scenes.

\begin{figure*}[t]
    \centering
    \includegraphics[width=\textwidth]{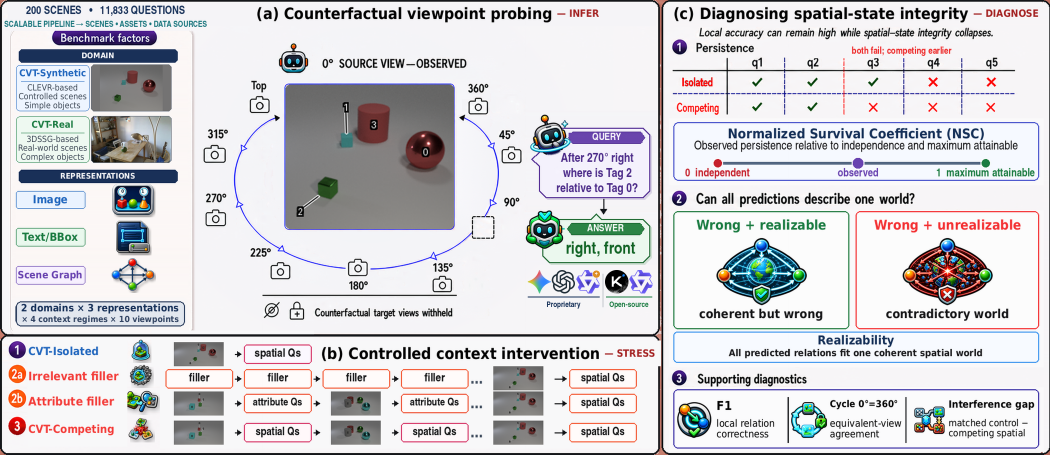}
    \caption{\textbf{CVT-Bench overview.}
    (a) From one observed source view, models answer relational queries under hidden counterfactual viewpoints across two domains and three input representations.
    (b) Controlled context interventions compare isolated evaluation, irrelevant filler, attribute filler, and competing spatial contexts. 
    (c) Complementary diagnostics measure local accuracy, relational persistence, equivalent-view consistency, global spatial realizability, and selective interference.
    }
    \label{fig:cvt_bench_overview}
\end{figure*}

To operationalize these principles, we introduce \textbf{CVT-Bench}. A model observes a scene from a single source view and answers relational queries across a sequence of unseen camera rotations, providing a computational analogue of Level-2 visual perspective taking in which target-view images are withheld. CVT-Bench evaluates spatial-state integrity through complementary measures of counterfactual accuracy, cycle consistency, the normalized survival coefficient, and spatial realizability. It pairs controlled synthetic scenes with annotated real-world 3DSSG environments and evaluates CVT-Isolated, CVT-Competing, irrelevant-filler, and attribute-filler contexts. Figure~\ref{fig:cvt_bench_overview} summarizes how these benchmark factors, matched interventions, and integrity diagnostics distinguish persistent spatial states from brittle, query-specific reasoning.

\definecolor{CVTBand}{HTML}{F2F6FB}
\definecolor{CVTOurs}{HTML}{EAF2FA}

\afterpage{\afterpage{%
\begin{table*}[t]
\centering
\scriptsize
\setlength{\tabcolsep}{3pt}
\renewcommand{\arraystretch}{1.12}
\begin{tabularx}{\textwidth}{
  >{\raggedright\arraybackslash}p{1.90cm}
  >{\raggedright\arraybackslash}X
  >{\raggedright\arraybackslash}X
  >{\raggedright\arraybackslash}X
  >{\raggedright\arraybackslash}X
  >{\raggedright\arraybackslash}X
}
\toprule
\textbf{Benchmark} &
  \textbf{Main focus} &
  \textbf{Visual evidence} &
  \textbf{Viewpoint probe} &
  \textbf{Context stress} &
  \textbf{Main diagnostic} \\
\midrule
\rowcolor{CVTBand}
SpinBench\newline\mbox{\cite{spinbench}}\newline{\color{black!55}\itshape ICLR'26} &
  Broad perspective-taking across 51 tasks &
  Task-specific visual inputs &
  Translation, rotation, and view selection &
  Independent instances &
  Task accuracy + augmentation consistency \\

SCOPE\newline\mbox{\cite{scope}}\newline{\color{black!55}\itshape ACL'26} &
  Cross-view spatial updating and integration &
  Single or up to eight observed views &
  Eight $45^\circ$ views; observed + imagined viewpoints &
  Single-scene multi-view &
  Accuracy + canonical-frame consistency \\

\rowcolor{CVTBand}
SpatialBench\newline\mbox{\cite{spatialbench}}\newline{\color{black!55}\itshape arXiv'25} &
  Hierarchical spatial cognition from observation to planning &
  Real-world first-person videos with calibrated LiDAR &
  Limited and task-specific &
  Video context across hierarchical tasks &
  Task accuracy + capability-oriented aggregate \\

MindCube\newline\mbox{\cite{mindcube}}\newline{\color{black!55}\itshape ICLR'26} &
  Spatial mental models, cognitive maps, and unseen-space inference &
  Limited real-world multi-view observations &
  Perspective taking and what-if motion beyond current view &
  Single-scene unseen-space inference &
  QA accuracy + map quality/isomorphism \\

\specialrule{\lightrulewidth}{0pt}{0pt}
\rowcolor{CVTOurs}
\textbf{CVT-Bench} &
  \textbf{Spatial-state integrity} across viewpoints and competing contexts &
  One source view as Image, Text/BBox, or Scene Graph; targets withheld &
  Counterfactual $0^\circ$--$360^\circ$ orbit + Top; fixed directed pairs &
  Isolated baseline vs irrelevant/attribute filler vs competing spatial contexts &
  F1; cycle consistency; NSC; realizability; interference gaps \\
\specialrule{\heavyrulewidth}{0pt}{0pt}
\end{tabularx}
\caption{\textbf{Diagnostic scope of CVT-Bench.}
Prior work emphasizes broad perspective taking, cross-view integration, hierarchical spatial cognition, or unseen-space inference. CVT-Bench targets spatial-state integrity by repeatedly probing fixed directed relations under hidden counterfactual viewpoints and competing scene contexts. Its diagnostics separate local accuracy, persistence, equivalent-view consistency, global realizability, and context-specific
interference.}
\label{tab:related_benchmarks}
\end{table*}
}}

Across five state-of-the-art MLLMs, CVT-Bench reveals a consistent diagnostic pattern. Models struggle especially when viewpoint changes exchange the left/right and front/behind axes, exhibit rapid decay in relational persistence, and lose global spatial realizability when multiple spatial contexts compete. 
These failures are not explained by generic context length or loss of scene facts alone because irrelevant filler restores much of the lost spatial accuracy and relational persistence, while attribute filler retrieval remains stable across prompt positions.
These weaknesses persist and intensify in real-world scenes, while targeted human studies in both domains confirm that the task is solvable. These results expose a gap between isolated spatial competence and robust spatial-state integrity.

We make four \textbf{contributions}: (1) we formalize \textit{spatial-state integrity} as a behavioral target for evaluating MLLMs beyond isolated accuracy; (2) we introduce \textit{CVT-Bench} and a scalable construction pipeline spanning controlled synthetic and annotated real-world scenes with counterfactual target views withheld; (3) we develop complementary measures of counterfactual accuracy, cycle consistency, the normalized survival coefficient, and spatial realizability, together with matched controls for context length and scene access; and (4) we identify a consistent bias toward simpler \(0^\circ/180^\circ\) relations and a selective collapse of spatial coherence under competing contexts.

\section{Related Work}

\textbf{Spatial Reasoning in Multimodal Models.}
Modern multimodal large language models (MLLMs), including GPT-5.2~\cite{gpt52report}, Gemini~\cite{team2023gemini}, Qwen-VL~\cite{wang2024qwen2}, Kosmos-2~\cite{peng2023kosmos}, and LLaVA~\cite{liu2023llava}, have demonstrated strong performance in visual question answering, grounding, and compositional reasoning~\cite{antol2015vqa,agrawal2019nocaps,xu2025mc}. A growing body of work further targets spatial competence through stronger visual grounding, structured representations, and spatial supervision. Groma~\cite{ma2024groma} and Ferret-v2~\cite{zhang2024ferret} improve region-level grounding; SpatialVLM~\cite{chen2024spatialvlm} introduces large-scale spatial supervision; and Struct2D~\cite{zhu2025struct2d} and LLaVA-ST~\cite{li2025llava} incorporate structured or spatial-temporal representations. These approaches primarily improve the extraction or prediction of spatial relations from observed inputs. Rather than introducing a new spatial reasoning model, we evaluate whether the relations produced by an MLLM remain compatible with a coherent spatial state across counterfactual viewpoints and contextual interference.

\textbf{Diagnostic Benchmarks for Spatial Cognition.}
General diagnostic benchmarks have shown that high aggregate multimodal accuracy can conceal systematic perceptual and reasoning failures. BLINK~\cite{fu2024blink}, MMVP~\cite{tong2024eyes}, and Winoground~\cite{thrush2022winoground} expose weaknesses in fine-grained perception, grounding, and compositional understanding. Spatially focused benchmarks extend this diagnostic perspective. SpatiaLQA~\cite{xie2026spatialqa} evaluates multi-step spatial logical reasoning in complex indoor scenes, while SpatialBench~\cite{spatialbench} organizes spatial cognition into a hierarchy spanning observation, relational understanding, symbolic reasoning, causal inference, and planning. Table~\ref{tab:related_benchmarks} summarizes the distinction: these benchmarks provide broad coverage across heterogeneous spatial tasks, whereas CVT-Bench asks whether same-scene responses remain transformable, persistent, and jointly realizable as viewpoint and surrounding context change.

\textbf{Perspective Taking and Cross-View Reasoning.}
Recent work most closely related to ours evaluates perspective taking and viewpoint-conditioned spatial reasoning. SpinBench~\cite{spinbench} decomposes perspective taking into progressively structured tasks involving translation, rotation, relative pose, and multi-object viewpoint change, revealing egocentric bias, rotational difficulty, and reformulation inconsistency. SCOPE~\cite{scope} evaluates spatial consistency, updating, and integration across observed and imagined viewpoints in synthetic and real-world scenes. LRR-Bench~\cite{lrrbench} similarly studies left/right judgments under object and camera rotation. As shown in Table~\ref{tab:related_benchmarks}, unlike task-competence or cross-view-consistency evaluations under task-specific visual evidence, CVT-Bench withholds target views and repeatedly probes one source-view spatial state across a controlled viewpoint sequence. It measures counterfactual accuracy, cycle consistency, the normalized survival coefficient, and spatial realizability under CVT-Isolated, irrelevant filler, attribute filler, and CVT-Competing contexts.

\textbf{Long-Context Consistency and Spatial Interference.}
LongBench~\cite{bai2024longbenchbilingualmultitaskbenchmark}, L-Eval~\cite{an2023levalinstitutingstandardizedevaluation}, RULER~\cite{hsieh2024rulerwhatsrealcontext}, and Lost in the Middle~\cite{liu2023lostmiddlelanguagemodels} demonstrate that model performance can degrade with context length and prompt position. Multi-image and interactive benchmarks similarly expose difficulties in integrating information and maintaining state over extended interactions~\cite{wang2024muirbenchcomprehensivebenchmarkrobust,liu2025agentbenchevaluatingllmsagents,yang2025aqabenchinteractivebenchmarkevaluating}. CVT-Bench instead uses shared context as a controlled intervention rather than a target capability: irrelevant filler controls token load and prompt position, while attribute filler tests scene access, distinguishing selective counterfactual degradation from generic context saturation or retrieval failure.

\section{Evaluating Spatial-State Integrity}


We view \textit{spatial-state integrity} as the behavioral capacity to preserve coherent scene predictions while transforming observer-relative relations across viewpoints and keeping scenes distinct. This formulation is inspired by Level-2 visual perspective taking, where the physical arrangement remains fixed while directional relations change with observer position. CVT-Bench uses reference-frame transformation as a computational probe, without assuming human-like mechanisms. Because all probes occur in one prompt and one joint response, persistence describes ordered response consistency rather than temporal memory or direct identification of a latent state.
A single correct response provides limited evidence and may instead reflect visual cues, linguistic priors, or a query-specific heuristic.
Counterfactual questions about the same scene are treated as jointly constrained probes of one inferred spatial state. 
As viewpoint changes, a model's predictions should persist across related transformations, agree at equivalent viewpoints, and collectively describe a possible spatial state. 

This perspective yields \textbf{four diagnostic dimensions}, namely transformation fidelity, relational persistence, global coherence, and contextual separability. They test whether inferred spatial states remain accurate, stable, realizable, and distinct across viewpoints and contexts.



\par\noindent\textbf{Transformation fidelity} measures whether observer-relative relations update correctly under hidden viewpoint changes, including cross-axis remapping at \(90^\circ/270^\circ\).

\par\noindent\textbf{Relational persistence} measures whether correct relations survive repeated viewpoint probes of the same scene.

\par\noindent\textbf{Global coherence} tests whether all predictions jointly admit a spatial configuration, separating coherent errors from contradictions.

\par\noindent\textbf{Contextual separability} tests whether one scene's spatial state remains distinct under competing scenes; matched fillers control for generic context and retrieval effects.

\par\noindent\textbf{Counterfactual Viewpoint Probing.} 
CVT-Bench presents a model with a single source-view representation of a scene and asks relational questions from a sequence of hypothetical observer viewpoints. The physical scene remains unchanged, but the observer is rotated about the vertical axis by 
$\theta {\in} \{0^\circ,45^\circ,90^\circ,135^\circ,180^\circ, 225^\circ,270^\circ,315^\circ,360^\circ\}$,
along with a top view (bird's-eye). No visual observation from a target viewpoint is provided. The model must therefore infer how the observer-relative relations among the same objects change under the specified transformation. 
Each query identifies an ordered object pair and asks which spatial relations hold under the requested viewpoint. Because multiple relations may hold simultaneously, such as \textit{front} and \textit{right}, the task is formulated as multi-label prediction. Queries at \(0^\circ\) measure spatial reasoning under direct observation, while counterfactual queries isolate the ability to transform relations without additional perceptual evidence. 
The controlled viewpoint orbit supports several complementary probes. Small rotations test whether relations that should remain unchanged are retained; \(180^\circ\) rotations reverse directions; \(90^\circ\) and \(270^\circ\) rotations exchange the left/right and front/behind axes; and \(0^\circ\) and \(360^\circ\) provide equivalent-view consistency checks.
\par\noindent\textbf{Input Representations.}
Each scene is evaluated through three input representations. Image provides the tagged source image and a tag-to-object dictionary without bounding boxes. Text/BBox removes the image and supplies object labels with 2D bounding boxes. Scene Graph augments the same Text/BBox representation with source-view relations among tagged objects. Counterfactual target views remain withheld in every mode.

\par\noindent\textbf{Question and Tracking Protocol.}
Questions use directed pairs of eligible visible or partially occluded objects. Up to three horizontal pairs are fixed across all ten viewpoints, two prioritizing partial occlusion when available; CVT-Real adds up to three above/below pairs. Other slots are sampled independently, except at \(360^\circ\), whose primary pairs and ground truth match \(0^\circ\) while phrasings differ. Because queries are shuffled, survival starts at each pair's first prompt occurrence and follows encounter order rather than angle or a fixed \(0^\circ\!\rightarrow360^\circ\) trajectory. Each scene contains up to 70 questions.

\par\noindent\textbf{Context Regimes.} 
CVT-Bench evaluates each scene under four context regimes designed to separate spatial-state interference from generic context effects.

\par\noindent\textbf{CVT-Isolated.}
A scene and its associated queries are presented without surrounding scene blocks. This condition measures counterfactual spatial reasoning in the absence of competing contexts and serves as the reference setting.

\par\noindent\textbf{CVT-Competing.}
Multiple independent scenes and their spatial queries are placed within the same prompt. Each scene remains geometrically unrelated to the others, and persistence is measured only across viewpoint probes belonging to the same scene. This tests whether spatial states can remain behaviorally separable within a shared context.
CVT-Competing uses 20 scenes, except GPT-5.2 uses 10 because 20-scene responses were output-truncated; its cross-model comparisons and aggregate results therefore reflect a lower interference load. 
Peak prompts occupy approximately 10--33\% of the advertised context windows, making input-context saturation unlikely; matched filler controls more directly test whether generic context load explains the degradation.

\par\noindent\textbf{Irrelevant filler.}
The other 19 scene blocks are replaced by a repeated irrelevant sentence matched to the removed text length, with no filler images or questions, while preserving the target scene's prompt position.

\par\noindent\textbf{Attribute filler.}
The other 19 scenes remain in context, but their spatial questions are replaced by low-complexity attribute-retrieval questions answerable directly from each scene dictionary. This preserves scene switching and visual load while removing competing spatial transformation.

\par\noindent\textbf{Models.} We evaluate Gemini 3.1 Pro, GPT-5.2, Kimi K2.5, Qwen 3.5 OS, and Qwen 3.5 Plus through public APIs at provider-default settings. Matched controls use the best-performing proprietary and open-weight models (Gemini 3.1 Pro and Qwen 3.5 OS), Image and Text/BBox modes, and target positions 0, 9, and 19. Results use one validated generation per job; exact settings and retry audits appear in the supplementary material.

\begin{figure*}
    \centering
    \includegraphics[width=1\linewidth]{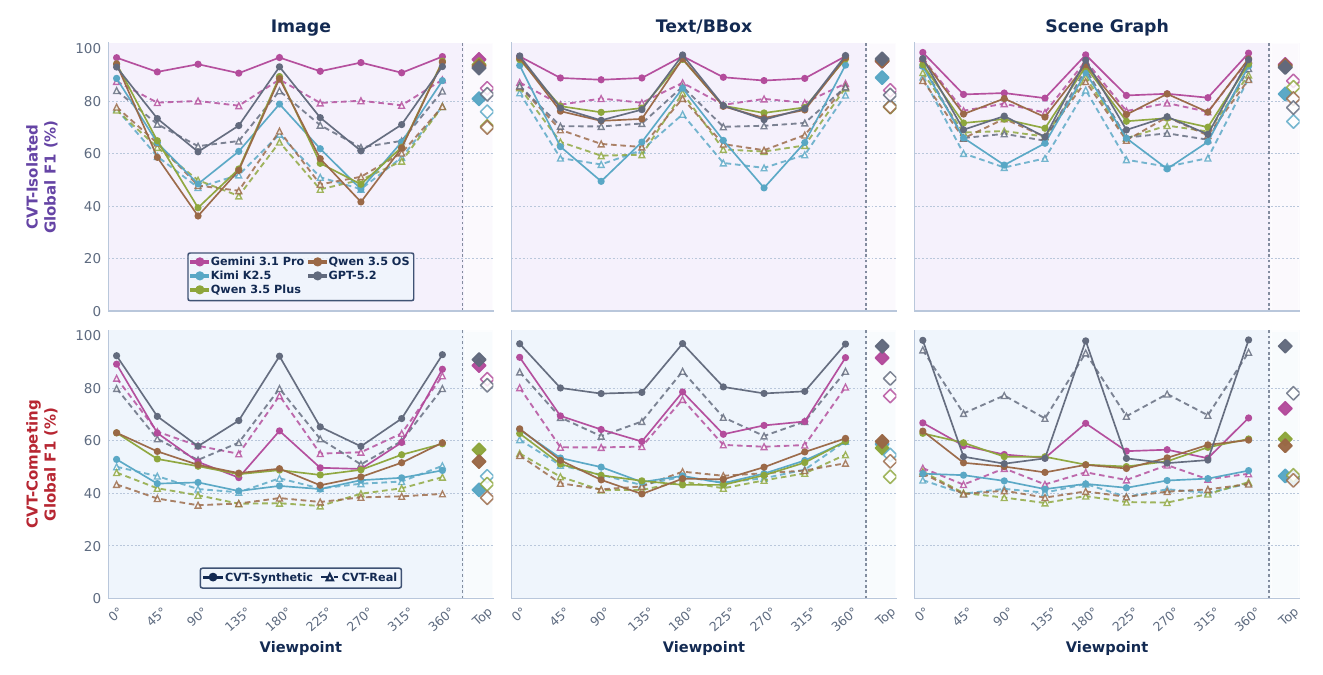}
\caption{\textbf{Global F1 across viewpoints.}
CVT-Isolated (top) and CVT-Competing (bottom) results under Image, Text/BBox, and Scene Graph inputs. GPT-5.2: 10 competing scenes; others: 20.}    \label{fig:viewpoint_f1}
\end{figure*}

\subsubsection{Evaluation Metrics}
Each metric corresponds to one diagnostic dimension of spatial-state integrity. We report viewpoint-conditioned multi-label F1 for transformation fidelity, the normalized survival coefficient for persistence, spatial realizability for global coherence, and interference gaps across context regimes for contextual separability. Cycle consistency is used as an additional equivalent-view diagnostic.
Ground-truth-dependent metrics use ambiguity-aware scoring that credits threshold-supported borderline relations, making reported scores upper bounds. Admissible alternatives vary only threshold-sensitive axes; opposite directions and unrelated axes never receive credit. We also confirm the same conclusions under strict scoring, as detailed in the supplementary material.

\par\noindent\textbf{Counterfactual accuracy.}
Let \(r_{n,\theta}\) and \(\hat r_{n,\theta}\) denote the ground-truth and predicted relation sets for query \(n\) at viewpoint \(\theta\). Since multiple directional predicates may hold simultaneously, we compute multi-label precision, recall, and F1. Global F1 includes every benchmark query, including \(0^\circ\), \(360^\circ\), and Top. We also separate the source-view score from all target viewpoints through the counterfactual gap:
\begin{equation}
\Delta_{\mathrm{CF}} = F1(0^\circ) - \frac{1}{|\Theta_{\mathrm{CF}}|} \sum_{\theta\in\Theta_{\mathrm{CF}}}F1(\theta),
\end{equation}
where \(\Theta_{\mathrm{CF}}\) contains every viewpoint except \(0^\circ\). A larger gap indicates poor transfer to unseen reference frames.

\par\noindent\textbf{Normalized survival coefficient.}
For tracked trajectory \(n\), let \(C_{n,t}=1\) indicate an exact prediction at sequence step \(t\), whose viewpoint is \(\theta_{n,t}\). For trajectories \(\mathcal T_k\) observed through step \(k\), prefix survival is
\begin{equation}
S_{\mathrm{obs}}(k)
=
\frac{1}{|\mathcal T_k|}
\sum_{n\in\mathcal T_k}
\prod_{t=1}^{k} C_{n,t}.
\end{equation}
Let \(p(v)\) denote exact tracked-pair accuracy at viewpoint \(v\), computed separately for each model, domain, context regime, and representation. Persistence expected under independent answering is
\begin{equation}
S_{\mathrm{ind}}(k)=
\frac{1}{|\mathcal T_k|}
\sum_{n\in\mathcal T_k}
\prod_{t=1}^{k}p(\theta_{n,t}).
\end{equation}
Let \(q_t\) be exact tracked-pair accuracy at survival step \(t\).
Maximum attainable prefix survival is \(S_{\mathrm{max}}(k)=\min_{1\leq t\leq k}q_t\).
We define
\begin{equation}
\mathrm{NSC}=
\frac{\sum_{k\in\mathcal K}[S_{\mathrm{obs}}(k)-S_{\mathrm{ind}}(k)]}
     {\sum_{k\in\mathcal K}[S_{\mathrm{max}}(k)-S_{\mathrm{ind}}(k)]}.
\end{equation}
Here, \(0\) denotes independent persistence, \(1\) maximum attainable persistence given marginal correctness, unclipped negative values indicate persistence below independence, and \(\mathcal K=\{2,4,6,8,10\}\). All terms use the same fixed tracked pairs.

\par\noindent\textbf{Spatial realizability.}
For each scene, we transform every prediction, including \(360^\circ\), Top, and CVT-Real vertical relations, to the canonical frame and express it as a linear inequality over one shared 2D (CVT-Synthetic) or 3D (CVT-Real) object-coordinate system. Let \(\hat{\mathcal R}_s\) denote this complete constraint set. We define
$\mathrm{Realizable}(s) = \mathbb{I} \left[ \exists\,\mathcal W \text{ such that } \mathcal W \models \hat{\mathcal R}_s \right]$,
where \(\mathcal W\) is a spatial configuration satisfying all predicted relations. We report the proportion of scenes whose predictions admit at least one such configuration, together with realizability conditioned on the prediction set containing an error. This separates coherent-but-incorrect spatial interpretations from mutually incompatible outputs.

\par\noindent\textbf{Spatial interference.}
For any metric \(M\), we define the interference gap between a matched control regime \(c\) and CVT-Competing as
$\Delta_{\mathrm{int}}^{M}(c) = M_c-M_{\mathrm{competing}}$.
We compute this gap using both irrelevant filler and attribute filler controls. Large gaps accompanied by stable attribute filler retrieval indicate selective degradation under competing spatial reasoning demands.

\par\noindent\textbf{Cycle consistency.}
Because \(0^\circ\) and \(360^\circ\) correspond to the same observer pose, we measure agreement over the \(N\) matched primary queries at these equivalent viewpoints:
$C_{\mathrm{cycle}} = \frac{1}{N} \sum_{n=1}^{N} \mathbb{I} \left[ \hat r_{n,0^\circ} = \hat r_{n,360^\circ} \right]$.
Cycle consistency measures equivalent-view agreement but does not establish correctness or global realizability.

\subsubsection{Benchmark Domains and Construction.} 
CVT-Bench pairs controlled CVT-Synthetic scenes, which isolate the target construct under exact geometry, with annotated CVT-Real scenes containing natural appearance, clutter, and occlusion. A shared pipeline generates questions and counterfactual ground truth from scene geometry and annotations, supporting new rendered assets and compatible annotated 3D sources.
Across both domains, sparse scenes contain 2--5 queried objects and dense scenes contain 6--10.

\begin{figure*}[t]
\centering
\begin{minipage}[c]{0.705\linewidth}
    \centering
    \includegraphics[width=\linewidth]{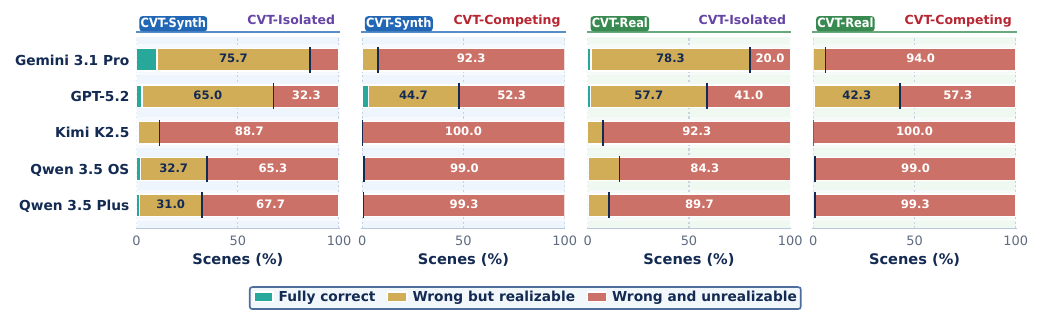}
\end{minipage}\hfill
\begin{minipage}[c]{0.285\linewidth}
    \centering
    {\small\bfseries Matched-control recovery}\par
    {\scriptsize
    \begin{tabular}{@{}cc@{}}
    I = Isolated & C = Competing \\
    IF = Irrelevant filler & AF = Attribute filler
    \end{tabular}}\par
    \vspace{3pt}
    \resizebox{\linewidth}{!}{%
    \begin{tabular}{@{}lrrrr@{}}
    \toprule
    \textbf{Model / metric} & \textbf{I} & \textbf{C} & \textbf{IF} & \textbf{AF} \\
    \midrule
    Gemini Image F1 & 95.5 & 69.5 & 88.3 & 67.7 \\
    Gemini Image NSC & .787 & .293 & .664 & .408 \\
    Qwen 3.5 OS Image F1 & 70.0 & 55.2 & 73.5 & 65.7 \\
    Qwen 3.5 OS Image NSC & .180 & $-.013$ & .200 & .179 \\
    \bottomrule
    \end{tabular}}
    \vspace{3pt}

    {\scriptsize\bfseries Attribute retrieval exact-match accuracy}\par
    {\scriptsize Gemini 93.0\% $\mid$ Qwen 3.5 OS 95.5\%}
\end{minipage}
\caption{\textbf{Global coherence and matched-control recovery.}
Left: ambiguity-aware fully correct, wrong but realizable, and wrong and unrealizable scene outcomes. Right: Image-mode F1/NSC across four matched regimes (15 scenes) and attribute-retrieval accuracy. GPT-5.2: 10 competing scenes; others: 20.}
\label{fig:realizability_by_model}
\end{figure*}

\textbf{Controlled synthetic scenes.}
For CVT-Synthetic, we generate 100 scenes and 5,703 questions using the CLEVR rendering pipeline~\cite{johnson2016clevrdiagnosticdatasetcompositional}, which provides exact object geometry, semantic annotations, and 3D coordinates. Objects are sampled from the standard CLEVR taxonomy and placed on a tabletop without stacking. Partial occlusions are retained, whereas objects that cannot be reliably identified are excluded from queries.
Tagged identifiers map to semantic descriptions, permitting concise, unambiguous references. On CVT-Synthetic, evaluated models correctly identify these tags, reducing object-recognition ambiguity as a confound.
Spatial relations are computed from ground-truth 3D coordinates in the corresponding observer frame and discretized into left, right, front, and behind predicates following the CLEVR protocol.

\textbf{Real-world scenes.}
To test external validity, CVT-Real contains 100 annotated 3DSSG indoor scenes~\cite{wald2020learning} and 6,130 questions. These scenes introduce natural object appearance, clutter, occlusion, and less regular spatial layouts while retaining the same protocol. The CVT-Real relation vocabulary is \(\mathcal R_{\mathrm{Real}}=\mathcal R_{\mathrm{Synthetic}}\cup\{\textit{above},\textit{below}\}\). Two annotators manually verified object identities and relations.

\section{Evaluation Results}

\par\noindent\textbf{Does Spatial Accuracy Change with Viewpoint?} 
Figure~\ref{fig:viewpoint_f1} shows accuracy peaking at the source-view condition, dropping most near \(90^\circ/270^\circ\), and partially recovering at opposite and equivalent views. Rotation difficulty is nonmonotonic because \(180^\circ\) reverses directions, whereas quarter turns exchange the left/right and front/behind axes. CVT-Competing amplifies intermediate-view errors across all modes. Across all non-source views, \(\Delta_{\mathrm{CF}}\) is .166/.139 on CVT-Synthetic and .151/.099 on CVT-Real under CVT-Isolated/CVT-Competing. Structured inputs help but do not remove this pattern, while cycle consistency falls from 99.1\% to 85.8\% on CVT-Synthetic and 97.9\% to 84.9\% on CVT-Real.

\begin{figure}[t]
    \centering
    \includegraphics[width=0.99\linewidth]{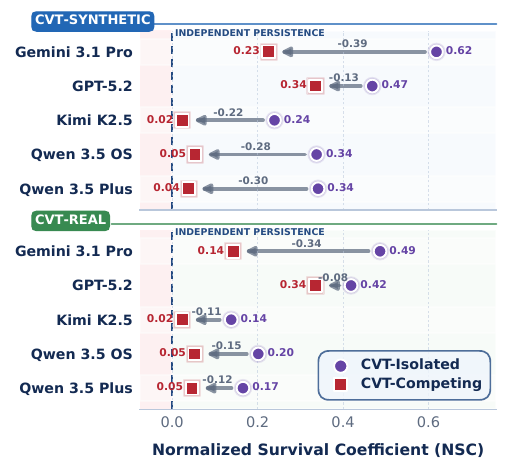}
\caption{\textbf{Normalized Survival Coefficient (NSC) across models.}
NSC locates observed persistence between independent answering (0) and maximum attainable persistence (1), aggregated over \(k\in\{2,4,6,8,10\}\) and averaged across representations. GPT-5.2: 10 competing scenes; others: 20.}
\label{fig:normalized_survival}
\end{figure}

\par\noindent\textbf{Does Correct Spatial Reasoning Persist across Viewpoints?}
Mean NSC falls from .401 under CVT-Isolated to .136 under CVT-Competing on CVT-Synthetic, and from .282 to .120 on CVT-Real, declining for every model and domain (Figure~\ref{fig:normalized_survival}). Even under CVT-Isolated, mean NSC remains below .5 in both domains, so competing contexts amplify an existing persistence deficit rather than create it. Thus, the collapse is neither model-specific nor confined to natural-scene complexity. Kimi and both Qwen variants approach zero under CVT-Competing, whereas GPT-5.2 retains more persistence under its smaller 10-scene load. Because NSC locates observed survival between independent answering and maximum attainable persistence, this decline cannot be reduced to lower pointwise accuracy alone. We also rule out unstable calculations, question difficulty, and encounter order as explanations, as detailed in the supplementary material.

\begin{table*}[t]
\centering
\scriptsize
\setlength{\tabcolsep}{3.0pt}
\renewcommand{\arraystretch}{1.08}
\begin{tabular*}{\textwidth}{@{\extracolsep{\fill}}lccccccc@{}}
\toprule
\multicolumn{2}{c}{\shortstack{\scriptsize\textit{Cell: Img/BB/SG}\\[-1pt]
\scriptsize\textbf{bold} = best mode per domain}} &
  \multicolumn{2}{c}{\textbf{Global F1 (\%, $\uparrow$)}} &
  \multicolumn{2}{c}{\textbf{NSC ($\uparrow$)}} &
  \multicolumn{2}{c}{\textbf{Realizability (\%, $\uparrow$)}} \\
\cmidrule(lr){1-2} \cmidrule(lr){3-4} \cmidrule(lr){5-6} \cmidrule(lr){7-8}
\textbf{Model} & \textbf{Domain} & \textbf{CVT-Isolated} & \textbf{CVT-Competing} &
\textbf{CVT-Isolated} & \textbf{CVT-Competing} &
\textbf{CVT-Isolated} & \textbf{CVT-Competing} \\
\midrule
Gemini 3.1 Pro & \begin{tabular}{@{}c@{}}\textbf{S} \\[-1.5pt] \textbf{R}\end{tabular} & \begin{tabular}{@{}c@{}}\textbf{94.2} / 92.5 / 89.6 \\[-1.5pt] 82.9 / 82.7 / \textbf{85.3}\end{tabular} & \begin{tabular}{@{}c@{}}68.2 / \textbf{76.7} / 61.6 \\[-1.5pt] \textbf{69.6} / 67.7 / 47.2\end{tabular} & \begin{tabular}{@{}c@{}}\textbf{0.64} / \textbf{0.64} / 0.58 \\[-1.5pt] 0.51 / \textbf{0.53} / 0.41\end{tabular} & \begin{tabular}{@{}c@{}}0.08 / \textbf{0.34} / 0.26 \\[-1.5pt] 0.08 / 0.11 / \textbf{0.24}\end{tabular} & \begin{tabular}{@{}c@{}}\textbf{97.0} / 83.0 / 77.0 \\[-1.5pt] \textbf{85.0} / \textbf{85.0} / 70.0\end{tabular} & \begin{tabular}{@{}c@{}}1.0 / \textbf{19.0} / 3.0 \\[-1.5pt] 4.0 / 6.0 / \textbf{8.0}\end{tabular} \\
\specialrule{0.25pt}{0pt}{0pt}
GPT-5.2 & \begin{tabular}{@{}c@{}}\textbf{S} \\[-1.5pt] \textbf{R}\end{tabular} & \begin{tabular}{@{}c@{}}80.3 / \textbf{86.1} / 82.9 \\[-1.5pt] 74.3 / 77.4 / \textbf{77.8}\end{tabular} & \begin{tabular}{@{}c@{}}77.8 / \textbf{87.6} / 76.1 \\[-1.5pt] 68.1 / 75.3 / \textbf{81.5}\end{tabular} & \begin{tabular}{@{}c@{}}\textbf{0.53} / \textbf{0.53} / 0.34 \\[-1.5pt] 0.43 / \textbf{0.48} / 0.35\end{tabular} & \begin{tabular}{@{}c@{}}0.35 / \textbf{0.57} / 0.08 \\[-1.5pt] 0.29 / \textbf{0.39} / 0.33\end{tabular} & \begin{tabular}{@{}c@{}}63.0 / \textbf{74.0} / 66.0 \\[-1.5pt] 55.0 / \textbf{73.0} / 49.0\end{tabular} & \begin{tabular}{@{}c@{}}38.0 / \textbf{64.0} / 41.0 \\[-1.5pt] 30.0 / \textbf{54.0} / 44.0\end{tabular} \\
\specialrule{0.25pt}{0pt}{0pt}
Kimi K2.5 & \begin{tabular}{@{}c@{}}\textbf{S} \\[-1.5pt] \textbf{R}\end{tabular} & \begin{tabular}{@{}c@{}}71.0 / 74.5 / \textbf{76.0} \\[-1.5pt] 62.9 / 68.4 / \textbf{70.4}\end{tabular} & \begin{tabular}{@{}c@{}}45.9 / \textbf{53.8} / 45.5 \\[-1.5pt] 45.7 / \textbf{51.3} / 42.4\end{tabular} & \begin{tabular}{@{}c@{}}0.24 / \textbf{0.26} / 0.22 \\[-1.5pt] 0.10 / \textbf{0.19} / 0.13\end{tabular} & \begin{tabular}{@{}c@{}}0.02 / 0.01 / \textbf{0.04} \\[-1.5pt] 0.01 / 0.01 / \textbf{0.05}\end{tabular} & \begin{tabular}{@{}c@{}}8.0 / \textbf{17.0} / 9.0 \\[-1.5pt] 5.0 / 8.0 / \textbf{10.0}\end{tabular} & \begin{tabular}{@{}c@{}}0.0 / 0.0 / 0.0 \\[-1.5pt] 0.0 / 0.0 / 0.0\end{tabular} \\
\specialrule{0.25pt}{0pt}{0pt}
Qwen 3.5 OS & \begin{tabular}{@{}c@{}}\textbf{S} \\[-1.5pt] \textbf{R}\end{tabular} & \begin{tabular}{@{}c@{}}71.8 / 85.2 / \textbf{85.9} \\[-1.5pt] 62.8 / 73.3 / \textbf{76.7}\end{tabular} & \begin{tabular}{@{}c@{}}53.5 / 53.8 / \textbf{55.7} \\[-1.5pt] 38.9 / \textbf{48.5} / 42.3\end{tabular} & \begin{tabular}{@{}c@{}}0.18 / \textbf{0.51} / 0.32 \\[-1.5pt] 0.06 / \textbf{0.28} / 0.27\end{tabular} & \begin{tabular}{@{}c@{}}0.01 / 0.05 / \textbf{0.11} \\[-1.5pt] 0.01 / 0.06 / \textbf{0.10}\end{tabular} & \begin{tabular}{@{}c@{}}9.0 / \textbf{51.0} / 44.0 \\[-1.5pt] 1.0 / 18.0 / \textbf{28.0}\end{tabular} & \begin{tabular}{@{}c@{}}0.0 / 0.0 / \textbf{3.0} \\[-1.5pt] 0.0 / 1.0 / \textbf{2.0}\end{tabular} \\
\specialrule{0.25pt}{0pt}{0pt}
Qwen 3.5 Plus & \begin{tabular}{@{}c@{}}\textbf{S} \\[-1.5pt] \textbf{R}\end{tabular} & \begin{tabular}{@{}c@{}}73.3 / \textbf{86.1} / 82.4 \\[-1.5pt] 61.6 / 71.7 / \textbf{77.8}\end{tabular} & \begin{tabular}{@{}c@{}}54.3 / 52.4 / \textbf{57.1} \\[-1.5pt] 41.6 / \textbf{47.4} / 41.5\end{tabular} & \begin{tabular}{@{}c@{}}0.20 / \textbf{0.52} / 0.30 \\[-1.5pt] 0.05 / 0.20 / \textbf{0.25}\end{tabular} & \begin{tabular}{@{}c@{}}0.02 / \textbf{0.05} / \textbf{0.05} \\[-1.5pt] 0.01 / 0.05 / \textbf{0.08}\end{tabular} & \begin{tabular}{@{}c@{}}7.0 / \textbf{53.0} / 37.0 \\[-1.5pt] 1.0 / 14.0 / \textbf{16.0}\end{tabular} & \begin{tabular}{@{}c@{}}0.0 / \textbf{1.0} / \textbf{1.0} \\[-1.5pt] 0.0 / \textbf{1.0} / \textbf{1.0}\end{tabular} \\
\bottomrule
\end{tabular*}
\caption{\textbf{Spatial-state integrity across domains, models, and representations.}
S/R = CVT-Synthetic/CVT-Real; bold compares modes within each model--domain cell, not models. GPT-5.2: 10 competing scenes; others: 20.}
\label{tab:combined_model_results}
\end{table*}

\par\noindent\textbf{Do MLLMs Form a Coherent Spatial World?}
Per-query accuracy does not reveal whether predictions across viewpoints jointly describe any possible spatial arrangement. A model may be wrong yet internally coherent, or it may produce mutually incompatible relations. We therefore evaluate \textit{spatial realizability}, which tests whether all viewpoint-conditioned predictions for a scene can be satisfied by a single spatial configuration after transformation into a common reference frame. Only asserted labels impose constraints, so omitted relations cannot make an output unrealizable. We also confirm ground-truth realizability, reduced realizability under deliberate errors, and connected query graphs, as detailed in the supplementary material.
Figure~\ref{fig:realizability_by_model} decomposes each scene-level output into ambiguity-aware \textit{fully correct}, \textit{wrong but realizable}, and \textit{wrong and unrealizable}. Under CVT-Isolated, 43.0\% of CVT-Synthetic scenes and 33.8\% of CVT-Real scenes are incorrect but coherent. Under CVT-Competing, these rates fall to 10.8\% and 10.0\%, while wrong-and-unrealizable outputs rise to 88.6\% and 89.9\%. Overall realizability drops from 46.3\% to 11.4\% on CVT-Synthetic and from 34.5\% to 10.1\% on CVT-Real. Thus, incoherence is already common in isolation and becomes dominant under competition, where interference converts coherent misinterpretations into relation sets that cannot describe any spatial world.
After matching model, representation, errors, and predicted relations, realizability still decreases by 16.3 points on CVT-Synthetic and 18.9 points on CVT-Real. A graded analysis shows the same decline; details appear in the supplementary material. Model profiles nevertheless differ, with Gemini showing the highest realizability under CVT-Isolated but dropping sharply, while Kimi and both Qwen variants approach zero under CVT-Competing. GPT-5.2's smaller decline reflects a lower 10-scene load and is not load-matched; Table~\ref{tab:combined_model_results} provides model-level results.

\par\noindent\textbf{Is the Coherence Collapse Caused by Generic Context Load?} 
Figure~\ref{fig:realizability_by_model} compares CVT-Competing with irrelevant and attribute filler. On the matched Image subset, irrelevant filler restores Gemini's F1 from 69.5 to 88.3 and NSC from .293 to .664; 
Qwen exhibits comparable recovery, with NSC increasing from $-0.013$ under CVT-Competing to $0.200$ with irrelevant filler.
Attribute filler gives incomplete spatial recovery, while retrieval remains 93.0\% and 95.5\%. Paired confidence intervals confirm irrelevant-filler F1 recovery for both models and attribute-filler F1 and NSC recovery for Qwen 3.5 OS; Gemini's attribute-filler recovery remains uncertain, with details in the supplementary material. These control-minus-competing differences define \(\Delta_{\mathrm{int}}\) and show that length, position, visual load, and simple retrieval are insufficient explanations alone, although the controls do not identify a latent mechanism.

\par\noindent\textbf{Why Are \(90^\circ\) and \(270^\circ\) Difficult?}
For incorrect two-relation answers at \(90^\circ/270^\circ\), we compare equal-error alternatives aligned or unaligned with the corresponding \(0^\circ/180^\circ\) relation. Under CVT-Isolated, models choose the aligned answer in 73.8--82.1\% of CVT-Synthetic errors, versus 4.4--7.9\% for the alternative, across all 15 model--representation combinations and after requiring a correct \(0^\circ\) answer. This preference for simpler \(0^\circ/180^\circ\) relations helps explain quarter-turn failures, without implying a particular internal algorithm.

\par\noindent\textbf{Does Spatial-State Integrity Generalize beyond Controlled Scenes?}
The same angular, persistence, and coherence patterns hold on CVT-Real despite natural appearance, clutter, occlusion, and irregular layouts. Mean F1 is 73.7 versus 82.1 under CVT-Isolated and 53.9 versus 61.3 under CVT-Competing (CVT-Real versus CVT-Synthetic). Identical questions across regimes control source-view observability, while exact-geometry CVT-Synthetic reproduces the trends. On CVT-Real, viewpoint-dependent horizontal relations lose NSC (.298 to .132), whereas invariant vertical relations remain persistent (.769 to .834) despite lower F1. Additionally, three annotators answered about 180 questions per domain across three diverse scenes, reaching F1 scores of 84.9\% and 81.8\% versus model averages of 79.9\% and 70.7\% on matched CVT-Isolated Image inputs. Although not ceiling estimates, these results support benchmark solvability and ground-truth validity.

\par\noindent\textbf{Do Structured Representations Improve Spatial-State Integrity?}
Table~\ref{tab:combined_model_results} shows that Text/BBox improves over Image in every domain and context regime, with averaged F1 rising by 4.9--6.8 points, NSC by 0.048--0.135, and realizability by 5.6--18.8 points. Scene Graph provides complementary gains and exceeds Image realizability in all domains and context regimes. Structured representations improve local accuracy, persistence, and coherence, but do not solve spatial-state instability.

\section{Conclusion and Limitations}
Across 200 scenes and 11,833 questions, CVT-Bench shows that local accuracy overstates spatial-state integrity. CVT-Competing reduces NSC and turns coherent errors into unrealizable predictions, while matched controls show that generic context load alone cannot explain the decline. Structured inputs help but do not eliminate instability. Our one-shot probe measures behavioral consistency, not temporal memory or latent mechanisms. Human studies cover only three scenes per domain, and CVT-Real is annotated rather than embodied. Future work should expand models, transformations, human evaluation, and embodied settings.

\bibliography{aaai2027}

@misc{johnson2016clevrdiagnosticdatasetcompositional,
      title={CLEVR: A Diagnostic Dataset for Compositional Language and Elementary Visual Reasoning}, 
      author={Justin Johnson and Bharath Hariharan and Laurens van der Maaten and Li Fei-Fei and C. Lawrence Zitnick and Ross Girshick},
      year={2016},
      eprint={1612.06890},
      archivePrefix={arXiv},
      primaryClass={cs.CV},
      url={https://arxiv.org/abs/1612.06890}, 
}

@misc{bai2024longbenchbilingualmultitaskbenchmark,
      title={LongBench: A Bilingual, Multitask Benchmark for Long Context Understanding}, 
      author={Yushi Bai and Xin Lv and Jiajie Zhang and Hongchang Lyu and Jiankai Tang and Zhidian Huang and Zhengxiao Du and Xiao Liu and Aohan Zeng and Lei Hou and Yuxiao Dong and Jie Tang and Juanzi Li},
      year={2024},
      eprint={2308.14508},
      archivePrefix={arXiv},
      primaryClass={cs.CL},
      url={https://arxiv.org/abs/2308.14508}, 
}

@misc{an2023levalinstitutingstandardizedevaluation,
      title={L-Eval: Instituting Standardized Evaluation for Long Context Language Models}, 
      author={Chenxin An and Shansan Gong and Ming Zhong and Xingjian Zhao and Mukai Li and Jun Zhang and Lingpeng Kong and Xipeng Qiu},
      year={2023},
      eprint={2307.11088},
      archivePrefix={arXiv},
      primaryClass={cs.CL},
      url={https://arxiv.org/abs/2307.11088}, 
}

@misc{hsieh2024rulerwhatsrealcontext,
      title={RULER: What's the Real Context Size of Your Long-Context Language Models?}, 
      author={Cheng-Ping Hsieh and Simeng Sun and Samuel Kriman and Shantanu Acharya and Dima Rekesh and Fei Jia and Yang Zhang and Boris Ginsburg},
      year={2024},
      eprint={2404.06654},
      archivePrefix={arXiv},
      primaryClass={cs.CL},
      url={https://arxiv.org/abs/2404.06654}, 
}

@misc{liu2023lostmiddlelanguagemodels,
      title={Lost in the Middle: How Language Models Use Long Contexts}, 
      author={Nelson F. Liu and Kevin Lin and John Hewitt and Ashwin Paranjape and Michele Bevilacqua and Fabio Petroni and Percy Liang},
      year={2023},
      eprint={2307.03172},
      archivePrefix={arXiv},
      primaryClass={cs.CL},
      url={https://arxiv.org/abs/2307.03172}, 
}

@misc{wang2024muirbenchcomprehensivebenchmarkrobust,
      title={MuirBench: A Comprehensive Benchmark for Robust Multi-image Understanding}, 
      author={Fei Wang and Xingyu Fu and James Y. Huang and Zekun Li and Qin Liu and Xiaogeng Liu and Mingyu Derek Ma and Nan Xu and Wenxuan Zhou and Kai Zhang and Tianyi Lorena Yan and Wenjie Jacky Mo and Hsiang-Hui Liu and Pan Lu and Chunyuan Li and Chaowei Xiao and Kai-Wei Chang and Dan Roth and Sheng Zhang and Hoifung Poon and Muhao Chen},
      year={2024},
      eprint={2406.09411},
      archivePrefix={arXiv},
      primaryClass={cs.CV},
      url={https://arxiv.org/abs/2406.09411}, 
}

@misc{liu2025agentbenchevaluatingllmsagents,
      title={AgentBench: Evaluating LLMs as Agents}, 
      author={Xiao Liu and Hao Yu and Hanchen Zhang and Yifan Xu and Xuanyu Lei and Hanyu Lai and Yu Gu and Hangliang Ding and Kaiwen Men and Kejuan Yang and Shudan Zhang and Xiang Deng and Aohan Zeng and Zhengxiao Du and Chenhui Zhang and Sheng Shen and Tianjun Zhang and Yu Su and Huan Sun and Minlie Huang and Yuxiao Dong and Jie Tang},
      year={2025},
      eprint={2308.03688},
      archivePrefix={arXiv},
      primaryClass={cs.AI},
      url={https://arxiv.org/abs/2308.03688}, 
}

@misc{yang2025aqabenchinteractivebenchmarkevaluating,
      title={AQA-Bench: An Interactive Benchmark for Evaluating LLMs' Sequential Reasoning Ability}, 
      author={Siwei Yang and Bingchen Zhao and Cihang Xie},
      year={2025},
      eprint={2402.09404},
      archivePrefix={arXiv},
      primaryClass={cs.CL},
      url={https://arxiv.org/abs/2402.09404}, 
}

@inproceedings{ma2024groma,
  title={Groma: Localized visual tokenization for grounding multimodal large language models},
  author={Ma, Chuofan and Jiang, Yi and Wu, Jiannan and Yuan, Zehuan and Qi, Xiaojuan},
  booktitle={European Conference on Computer Vision},
  pages={417--435},
  year={2024},
  organization={Springer}
}

@inproceedings{li2025llava,
  title={Llava-st: A multimodal large language model for fine-grained spatial-temporal understanding},
  author={Li, Hongyu and Chen, Jinyu and Wei, Ziyu and Huang, Shaofei and Hui, Tianrui and Gao, Jialin and Wei, Xiaoming and Liu, Si},
  booktitle={Proceedings of the IEEE/CVF Conference on Computer Vision and Pattern Recognition},
  pages={8592--8603},
  year={2025}
}

@inproceedings{liu2023llava,
  title={Visual Instruction Tuning},
  author={Liu, Haotian and Li, Chunyuan and Wu, Qingyang and Lee, Yong Jae},
  booktitle={Advances in Neural Information Processing Systems},
  volume={36},
  year={2023}
}

@inproceedings{wald2020learning,
  title={Learning 3D Semantic Scene Graphs From 3D Indoor Reconstructions},
  author={Wald, Johanna and Dhamo, Helisa and Navab, Nassir and Tombari, Federico},
  booktitle={Proceedings of the IEEE/CVF Conference on Computer Vision and Pattern Recognition},
  pages={3961--3970},
  year={2020}
}

@article{zhu2025struct2d,
  title={Struct2D: A Perception-Guided Framework for Spatial Reasoning in MLLMs},
  author={Zhu, Fangrui and Wang, Hanhui and Xie, Yiming and Gu, Jing and Ding, Tianye and Yang, Jianwei and Jiang, Huaizu},
  journal={arXiv preprint arXiv:2506.04220},
  year={2025}
}

@article{xie2026spatialqa,
  title={SpatiaLQA: A Benchmark for Evaluating Spatial Logical Reasoning in Vision-Language Models},
  author={Xie, Yuechen and Zhang, Xiaoyan and Shan, Yicheng and Zhu, Hao and Tang, Rui and Wei, Rong and Song, Mingli and Wan, Yuanyu and Song, Jie},
  journal={arXiv preprint arXiv:2602.20901},
  year={2026}
}

@article{zhang2024ferret,
  title={Ferret-v2: An improved baseline for referring and grounding with large language models},
  author={Zhang, Haotian and You, Haoxuan and Dufter, Philipp and Zhang, Bowen and Chen, Chen and Chen, Hong-You and Fu, Tsu-Jui and Wang, William Yang and Chang, Shih-Fu and Gan, Zhe and others},
  journal={arXiv preprint arXiv:2404.07973},
  year={2024}
}

@inproceedings{chen2024spatialvlm,
  title={Spatialvlm: Endowing vision-language models with spatial reasoning capabilities},
  author={Chen, Boyuan and Xu, Zhuo and Kirmani, Sean and Ichter, Brain and Sadigh, Dorsa and Guibas, Leonidas and Xia, Fei},
  booktitle={Proceedings of the IEEE/CVF Conference on Computer Vision and Pattern Recognition},
  pages={14455--14465},
  year={2024}
}

@inproceedings{thrush2022winoground,
  title={Winoground: Probing vision and language models for visio-linguistic compositionality},
  author={Thrush, Tristan and Jiang, Ryan and Bartolo, Max and Singh, Amanpreet and Williams, Adina and Kiela, Douwe and Ross, Candace},
  booktitle={Proceedings of the IEEE/CVF Conference on Computer Vision and Pattern Recognition},
  pages={5238--5248},
  year={2022}
}

@inproceedings{tong2024eyes,
  title={Eyes wide shut? exploring the visual shortcomings of multimodal llms},
  author={Tong, Shengbang and Liu, Zhuang and Zhai, Yuexiang and Ma, Yi and LeCun, Yann and Xie, Saining},
  booktitle={Proceedings of the IEEE/CVF Conference on Computer Vision and pattern recognition},
  pages={9568--9578},
  year={2024}
}

@inproceedings{fu2024blink,
  title={Blink: Multimodal large language models can see but not perceive},
  author={Fu, Xingyu and Hu, Yushi and Li, Bangzheng and Feng, Yu and Wang, Haoyu and Lin, Xudong and Roth, Dan and Smith, Noah A and Ma, Wei-Chiu and Krishna, Ranjay},
  booktitle={European Conference on Computer Vision},
  pages={148--166},
  year={2024},
  organization={Springer}
}

@article{michelon2006two,
  title={Two kinds of visual perspective taking},
  author={Michelon, Pascale and Zacks, Jeffrey M},
  journal={Perception \& psychophysics},
  volume={68},
  number={2},
  pages={327--337},
  year={2006},
  doi={10.3758/BF03193680},
  publisher={Springer}
}

@inproceedings{spinbench,
  title={{SpinBench}: Perspective and Rotation as a Lens on Spatial Reasoning in {VLMs}},
  author={Zhang, Yuyou and Corcodel, Radu and Hori, Chiori and Cherian, Anoop and Zhao, Ding},
  booktitle={International Conference on Learning Representations},
  year={2026},
  url={https://iclr.cc/virtual/2026/poster/10007175}
}

@inproceedings{scope,
  title={Diagnosing Spatial Consistency across Perspectives and Viewpoints in Large Vision-Language Models},
  author={Kim, Yoonji and Kim, Jieun and Jeong, Yujin and Cho, Sung-Bae},
  booktitle={Proceedings of the 64th Annual Meeting of the Association for Computational Linguistics (Volume 1: Long Papers)},
  pages={32803--32827},
  year={2026},
  publisher={Association for Computational Linguistics},
  doi={10.18653/v1/2026.acl-long.1514},
  url={https://aclanthology.org/2026.acl-long.1514/}
}

@misc{spatialbench,
  title={{SpatialBench}: Benchmarking Multimodal Large Language Models for Spatial Cognition},
  author={Xu, Peiran and Wang, Sudong and Zhu, Yao and Li, Jianing and Zhang, Yunjian},
  year={2025},
  eprint={2511.21471},
  archivePrefix={arXiv},
  primaryClass={cs.CV},
  url={https://arxiv.org/abs/2511.21471}
}

@inproceedings{mindcube,
  title={{MindCube}: Spatial Mental Modeling from Limited Views},
  author={Wang, Qineng and Yin, Baiqiao and Zhang, Pingyue and Zhang, Jianshu and Wang, Kangrui and Wang, Zihan and Zhang, Jieyu and Chandrasegaran, Keshigeyan and Liu, Han and Krishna, Ranjay and Xie, Saining and Wu, Jiajun and Fei-Fei, Li and Li, Manling},
  booktitle={International Conference on Learning Representations},
  year={2026},
  url={https://iclr.cc/virtual/2026/poster/10011954}
}

@misc{lrrbench,
  title={{LRR-Bench}: Left, Right or Rotate? Vision-Language Models Still Struggle With Spatial Understanding Tasks},
  author={Kong, Fei and Duan, Jinhao and Xu, Kaidi and Guo, Zhenhua and Zhu, Xiaofeng and Shi, Xiaoshuang},
  year={2025},
  eprint={2507.20174},
  archivePrefix={arXiv},
  primaryClass={cs.CV},
  url={https://arxiv.org/abs/2507.20174}
}

@article{team2023gemini,
  title={Gemini: a family of highly capable multimodal models},
  author={Team, Gemini and Anil, Rohan and Borgeaud, Sebastian and Alayrac, Jean-Baptiste and Yu, Jiahui and Soricut, Radu and Schalkwyk, Johan and Dai, Andrew M and Hauth, Anja and Millican, Katie and others},
  journal={arXiv preprint arXiv:2312.11805},
  year={2023}
}

@article{wang2024qwen2,
  title={Qwen2-vl: Enhancing vision-language model's perception of the world at any resolution},
  author={Wang, Peng and Bai, Shuai and Tan, Sinan and Wang, Shijie and Fan, Zhihao and Bai, Jinze and Chen, Keqin and Liu, Xuejing and Wang, Jialin and Ge, Wenbin and others},
  journal={arXiv preprint arXiv:2409.12191},
  year={2024}
}

@techreport{gpt52report,
  author      = {OpenAI},
  title       = {Update to {GPT}-5 System Card: {GPT}-5.2},
  institution = {OpenAI},
  year        = {2025},
  month       = {December},
  url         = {https://cdn.openai.com/pdf/3a4153c8-c748-4b71-8e31-aecbde944f8d/oai_5_2_system-card.pdf},
  note        = {Accessed: 2026-03-03}
}

@inproceedings{antol2015vqa,
  title={Vqa: Visual question answering},
  author={Antol, Stanislaw and Agrawal, Aishwarya and Lu, Jiasen and Mitchell, Margaret and Batra, Dhruv and Zitnick, C Lawrence and Parikh, Devi},
  booktitle={Proceedings of the IEEE International Conference on Computer Vision},
  pages={2425--2433},
  year={2015}
}

@article{peng2023kosmos,
  title={Kosmos-2: Grounding multimodal large language models to the world},
  author={Peng, Zhiliang and Wang, Wenhui and Dong, Li and Hao, Yaru and Huang, Shaohan and Ma, Shuming and Wei, Furu},
  journal={arXiv preprint arXiv:2306.14824},
  year={2023}
}

@inproceedings{xu2025mc,
  title={Mc-bench: A benchmark for multi-context visual grounding in the era of mllms},
  author={Xu, Yunqiu and Zhu, Linchao and Yang, Yi},
  booktitle={Proceedings of the IEEE/CVF International Conference on Computer Vision},
  pages={17675--17687},
  year={2025}
}

@inproceedings{agrawal2019nocaps,
  title={Nocaps: Novel object captioning at scale},
  author={Agrawal, Harsh and Desai, Karan and Wang, Yufei and Chen, Xinlei and Jain, Rishabh and Johnson, Mark and Batra, Dhruv and Parikh, Devi and Lee, Stefan and Anderson, Peter},
  booktitle={Proceedings of the IEEE/CVF international conference on computer vision},
  pages={8948--8957},
  year={2019}
}


\clearpage
\twocolumn[{
\centering
{\LARGE\bfseries Appendix\par}
\vspace{1.5em}
}]
\setcounter{secnumdepth}{2}
\setcounter{figure}{0}
\setcounter{table}{0}
\renewcommand{\thefigure}{S\arabic{figure}}
\renewcommand{\thetable}{S\arabic{table}}

This supplement documents benchmark construction, complete results, metric validation,
matched controls, qualitative examples, prompts, and reproducibility details. All reported
numbers use the same frozen benchmark, ambiguity-aware evaluation, and canonical predictions
as the main paper. We use \textbf{CVT-Isolated} and \textbf{CVT-Competing} throughout.

\section{Benchmark Construction and Composition}

\subsection{Factorial configuration}

CVT-Bench contains 100 CVT-Synthetic scenes and 100 CVT-Real scenes. Each scene is
evaluated through Image, Text/BBox, and Scene Graph representations under CVT-Isolated
and CVT-Competing. Matched irrelevant-filler and attribute-filler controls are additionally
evaluated on Gemini 3.1 Pro and Qwen 3.5 OS in Image and Text/BBox modes.

Ten viewpoint conditions are used: nine azimuths from \(0^\circ\) through \(360^\circ\)
at \(45^\circ\) increments, plus Top. Only the source view is observed.
Every counterfactual target view remains hidden. A fixed directed pair recurs across
viewpoints, and persistence follows the shuffled order in which its questions occur in the
prompt rather than numerical angle order.

\begin{table*}[t]
\centering
\small
\setlength{\tabcolsep}{4pt}
\renewcommand{\arraystretch}{1.10}
\begin{tabularx}{\textwidth}{lrrYY}
\toprule
\textbf{Property} & \textbf{CVT-Synthetic} & \textbf{CVT-Real} &
\textbf{Synthetic detail} & \textbf{Real detail} \\
\midrule
Scenes & 100 & 100 & 50 sparse / 50 dense & 54 sparse / 46 dense \\
Objects per scene & 2--10 (6.15 avg.) & 3--10 (5.27 avg.) & Controlled CLEVR objects & Natural 3DSSG objects \\
Questions & 5,703 & 6,130 & 20--60 (57.03 avg.) & 40--69 (61.30 avg.) \\
Source-view questions & 1,284 & 1,234 & \(0^\circ\) & \(0^\circ\), including vertical enrichment \\
Counterfactual questions & 4,419 & 4,896 & 491 per target condition & 544 per target condition \\
Fixed tracking pairs & 297 & 494 & 297 horizontal & 300 horizontal + 194 vertical-enriched \\
Fixed question instances & 2,970 & 4,940 & Ten appearances per pair & Ten appearances per pair \\
Scenes with partial occlusion & 43 & 38 & 1,315 affected questions & 1,268 affected questions \\
Multi-label questions & 5,265 & 4,782 & 10,968 labels total & 12,478 labels total \\
Relation vocabulary & 4 & 6 & left/right/front/behind & Synthetic set \(\cup\) above/below \\
Input resolution & \(320{\times}240\) & \(216{\times}384\) or \(384{\times}216\) & Fixed render size & Aspect-preserving canonical resize \\
\bottomrule
\end{tabularx}
\caption{\textbf{Frozen benchmark composition.} Counts are computed directly from canonical benchmark JSON files.}
\label{tab:supp_composition}
\end{table*}
\subsection{CVT-Synthetic construction}

CVT-Synthetic retains the frozen CLEVR-based protocol. Sparse scenes initially contain
3--5 objects and dense scenes 7--10. Objects with at least 5\% estimated visible area are
tagged; objects with at most 70\% visible area are marked partially occluded. Candidate
relations are derived from Blender coordinates with a 0.2-world-unit stability margin.
Up to two fixed pairs
must involve at least one partially occluded object when such pairs exist, and the fixed set
is completed to three pairs. Two additional pairs fill the five primary slots at each view;
up to ten more source-view pairs fill supplemental slots. At \(360^\circ\), the five primary
pairs exactly match \(0^\circ\). The final question order is shuffled after construction.

Image mode supplies the tagged source image and tag-to-object dictionary. Text/BBox supplies
the same dictionary with \([x,y,w,h]\) boxes and no pixels. Scene Graph supplies the Text/BBox
dictionary plus all original-view directed relations among tagged objects. Thus, Scene Graph
contains bounding boxes as well as relational structure.

\subsection{CVT-Real construction}

CVT-Real adapts the identical question and tracking protocol to 3DSSG scenes. Metadata is
authoritative for tag identity, object label, full projected bounding box, camera basis, and
3D object geometry. Tagged model inputs are canonical \(216{\times}384\) portrait or
\(384{\times}216\) landscape images. The same 5\% visible-area and 70\% partial-occlusion
thresholds are applied using union boxes and depth ordering; at least 350 projected pixels
are required by the visibility audit.

Only original-view pairs are filtered for ambiguity. The conservative gate requires
12-pixel/0.15-normalized lateral and 0.20 normalized depth separation; an
8-pixel/0.10-lateral/0.15-depth fallback is used only until three unordered pairs exist.
Admitted pairs propagate to every target view without view-specific filtering.

CVT-Real adds above/below when oriented 3D boxes and image evidence agree. Clean vertical
pairs require at most 20\% overlap of the smaller height interval, at least 12 pixels and
0.20 normalized vertical center separation, and consistent world-height/image ordering.
Up to five multi-axis vertical questions replace supplemental source-view slots, and up to
three unique pairs become trackers; inverse questions never inflate support. Twenty-three
scenes have no vertical tracker, 10 have one, 17 have two, and 50 have three (194 total).

\subsection{Question counts and regime assembly}

Each source view contains up to 15 horizontal questions, with CVT-Real vertical questions
occupying existing supplemental capacity. Every other view contains up to five horizontal
questions; a tracked vertical-enriched pair may add one question, producing up to 70 questions
per real scene. CVT-Isolated contains one scene per API request. CVT-Competing interleaves
complete scene blocks within one request:

\begin{center}
\texttt{scene/image 1} \(\rightarrow\) \texttt{its questions} \(\rightarrow\)
\texttt{scene/image 2} \(\rightarrow\) \texttt{its questions} \(\rightarrow\cdots\)
\end{center}

Gemini, Kimi, and both Qwen variants use 20 scenes per competing prompt. GPT-5.2 uses 10
because reasoning-enabled 20-scene responses were repeatedly output-truncated. We record this as
a distinct C20 completion/scalability failure rather than converting incomplete outputs into spatial
errors; all GPT-5.2 spatial metrics therefore use complete C10 responses, with the lower interference
load stated beside every model-level comparison. Numeric scene order is frozen sparse-first and
dense-last, with only one mixed-density block.

\FloatBarrier
\section{Metrics and Validation}

\subsection{Ambiguity-aware F1}

Global multi-label F1 includes all questions, including \(0^\circ\), \(360^\circ\), and Top.
Canonical ground truth remains unchanged. A separate frozen sidecar enumerates only
threshold-supported alternatives near relation boundaries. The reported upper bound credits
one of these alternatives, while strict scoring accepts only the canonical set. Opposite
directions and unrelated-axis substitutions never receive credit.

Strict and ambiguity-aware F1 differ by only .0040/.0046 on CVT-Synthetic and
.0094/.0104 on CVT-Real under CVT-Isolated/CVT-Competing. The central conclusions therefore
do not depend on ambiguity handling.

\subsection{Normalized survival coefficient}

For fixed tracked trajectory \(n\), let \(C_{n,t}\in\{0,1\}\) indicate an exact, ambiguity-aware
relation-set match at encounter step \(t\). For trajectories eligible at prefix \(k\),

\begin{equation}
S_{\mathrm{obs}}(k)=\frac{1}{|\mathcal T_k|}\sum_{n\in\mathcal T_k}\prod_{t=1}^{k}C_{n,t}.
\end{equation}

Let \(p(v)\) denote exact tracked-pair accuracy at viewpoint \(v\), computed separately for
each model, domain, context regime, and representation. The independent-answering reference is

\begin{equation}
S_{\mathrm{ind}}(k)=\frac{1}{|\mathcal T_k|}\sum_{n\in\mathcal T_k}
\prod_{t=1}^{k}p(\theta_{n,t}).
\end{equation}

Let \(q_t\) be exact accuracy at encounter step \(t\) over the same eligible trajectories.
The maximum possible prefix intersection given these marginals is
\(S_{\max}(k)=\min_{1\leq t\leq k}q_t\). We aggregate \(K=\{2,4,6,8,10\}\):

\begin{equation}
\mathrm{NSC}=
\frac{\sum_{k\in K}[S_{\mathrm{obs}}(k)-S_{\mathrm{ind}}(k)]}
{\sum_{k\in K}[S_{\max}(k)-S_{\mathrm{ind}}(k)]}.
\end{equation}

NSC \(=0\) represents independent answering and NSC \(=1\) maximum attainable persistence;
negative values are retained. Across all 60 model--domain--regime--representation cells, the
denominator ranges from .364 to 2.503 (median 1.642). All 60 denominators and all 300
constituent step gaps are positive; the smallest gap is .068.

\subsection{Spatial realizability}

Every asserted prediction is transformed to one canonical coordinate frame. Each scene assigns
a shared coordinate \(x_i\in\mathbb R^2\) in CVT-Synthetic or \(x_i\in\mathbb R^3\) in
CVT-Real to every queried object. For subject \(i\), reference \(j\), predicted relation \(r\),
and canonical direction \(d(\theta,r)\), the solver enforces

\begin{equation}
d(\theta,r)^\top(x_i-x_j)\geq1.
\end{equation}

Yaw rotates horizontal directions into the source frame; \(360^\circ\) equals \(0^\circ\);
Top uses the audited ground-plane mapping; above/below are invariant. Every label in a
multi-label answer adds a constraint. Omitted labels add none, making infeasibility conservative.
Translation is fixed by pinning one object to the origin; the unit margin is without loss of
generality because coordinates may scale. Linear-program feasibility defines binary
realizability. A mixed-integer program also maximizes the simultaneously satisfiable fraction
(MaxSAT), providing a graded diagnostic.

\begin{table*}[t]
\centering
\small
\setlength{\tabcolsep}{5pt}
\renewcommand{\arraystretch}{1.08}
\begin{tabular}{llrrrr}
\toprule
\textbf{Domain} & \textbf{Validation} & \textbf{Correct map} & \textbf{Opposite map} &
\textbf{Connected} & \textbf{Positive cycle rank} \\
\midrule
CVT-Synthetic & Ground-truth constraints & 100.0 & 0.0 & 100.0 & 97.0 \\
CVT-Real & Ground-truth constraints & 100.0 & 0.0 & 100.0 & 93.0 \\
\midrule
\multicolumn{2}{l}{\textbf{Perturbation / levels}} & \textbf{Level 1} & \textbf{Level 2} & \textbf{Level 3} & \textbf{Level 4} \\
\midrule
CVT-Synthetic & Relation flip (1/2/5/10\%) & 36.4 & 12.6 & 1.0 & 0.1 \\
CVT-Real & Relation flip (1/2/5/10\%) & 30.1 & 10.5 & 0.5 & 0.0 \\
CVT-Synthetic & Pair permutation (5/10/25/50\%) & 48.3 & 13.2 & 0.4 & 0.0 \\
CVT-Real & Pair permutation (5/10/25/50\%) & 45.4 & 12.4 & 0.1 & 0.0 \\
\bottomrule
\end{tabular}
\caption{\textbf{Realizability validation (\%).} Correct ground truth is universally feasible,
the wrong camera convention is universally infeasible, and sparse random corruption sharply
reduces realizability. A complete label inversion can remain coherent by symmetry, confirming
that realizability measures internal compatibility rather than correctness.}
\label{tab:supp_realizability_validation}
\end{table*}
The median CVT-Synthetic constraint system contains six objects, 114 inequalities, 20.5
directed pairs, 13 undirected pairs, nine repeated directed pairs, and cycle rank eight.
CVT-Real medians are five objects, 109 inequalities, 13 directed pairs, 8.5 undirected pairs,
six repeated directed pairs, and cycle rank four. Thus, the test is supported by overlapping,
connected query graphs rather than isolated pair checks.

\subsection{Cycle consistency and context gaps}

Cycle consistency is exact agreement between model predictions for matched \(0^\circ\) and
\(360^\circ\) questions. The counterfactual gap is \(0^\circ\) F1 minus mean F1 over all nine
non-source conditions. For metric \(M\), the interference gap is
\(\Delta_{\mathrm{int}}(c)=M(c)-M(\mathrm{CVT\mbox{-}Competing})\), where \(c\) is a matched
control. These diagnostics respectively isolate equivalent-view agreement, target-view loss,
and context-specific disruption.

\FloatBarrier
\section{Complete and Extended Results}

\subsection{All model, domain, regime, and representation cells}

Table~\ref{tab:supp_all_results} expands the main-paper aggregates into every evaluated cell.
Competition reduces F1, NSC, and realizability in both domains and almost every
model--representation combination. The decline is therefore not attributable to one model,
one input format, or one scene source. CVT-Real is additionally harder under both regimes,
especially for persistence and binary coherence, despite retaining substantial local F1.

Structured evidence helps without eliminating the instability. Averaged within each
domain--regime block, Text/BBox exceeds Image by 4.9--6.8 F1 points, .048--.135 NSC, and
5.6--18.8 realizability points. Scene Graph also exceeds Image realizability in all four blocks.
These row-wise trends motivate structured inputs as transparent baselines rather than as new
model rankings; competing prompt size differs for GPT-5.2 as stated in the caption.

\begin{figure*}[t]
\centering
\scriptsize
\setlength{\tabcolsep}{2.5pt}
\renewcommand{\arraystretch}{1.04}
\resizebox{\textwidth}{!}{%
\begin{tabular}{lllccc|ccc|ccc}
\toprule
&&&\multicolumn{3}{c|}{\textbf{F1}}&\multicolumn{3}{c|}{\textbf{NSC}}&\multicolumn{3}{c}{\textbf{Realizability}}\\
\textbf{Model}&\textbf{Domain}&\textbf{Regime}&\textbf{Img}&\textbf{BB}&\textbf{SG}&
\textbf{Img}&\textbf{BB}&\textbf{SG}&\textbf{Img}&\textbf{BB}&\textbf{SG}\\
\midrule
Gemini 3.1 Pro&S&I&.942&.925&.896&.639&.640&.577&.970&.830&.770\\
GPT-5.2&S&I&.803&.861&.829&.534&.532&.339&.630&.740&.660\\
Kimi K2.5&S&I&.710&.745&.760&.235&.260&.223&.080&.170&.090\\
Qwen 3.5 OS&S&I&.718&.852&.859&.182&.511&.322&.090&.510&.440\\
Qwen 3.5 Plus&S&I&.733&.861&.824&.199&.523&.302&.070&.530&.370\\
\rowcolor{CVTPaleBlue}
Gemini 3.1 Pro&S&C&.682&.767&.616&.081&.341&.256&.010&.190&.030\\
\rowcolor{CVTPaleBlue}
GPT-5.2&S&C&.778&.876&.761&.353&.570&.085&.380&.640&.410\\
\rowcolor{CVTPaleBlue}
Kimi K2.5&S&C&.459&.538&.455&.017&.013&.040&.000&.000&.000\\
\rowcolor{CVTPaleBlue}
Qwen 3.5 OS&S&C&.535&.538&.557&.008&.048&.106&.000&.000&.030\\
\rowcolor{CVTPaleBlue}
Qwen 3.5 Plus&S&C&.543&.524&.571&.018&.049&.050&.000&.010&.010\\
\midrule
Gemini 3.1 Pro&R&I&.829&.827&.853&.513&.532&.414&.850&.850&.700\\
GPT-5.2&R&I&.743&.774&.778&.428&.482&.347&.550&.730&.490\\
Kimi K2.5&R&I&.629&.684&.704&.097&.186&.132&.050&.080&.100\\
Qwen 3.5 OS&R&I&.628&.733&.767&.061&.278&.265&.010&.180&.280\\
Qwen 3.5 Plus&R&I&.616&.717&.778&.051&.196&.250&.010&.140&.160\\
\rowcolor{CVTPaleGreen}
Gemini 3.1 Pro&R&C&.696&.677&.472&.075&.113&.240&.040&.060&.080\\
\rowcolor{CVTPaleGreen}
GPT-5.2&R&C&.681&.753&.815&.285&.394&.326&.300&.540&.440\\
\rowcolor{CVTPaleGreen}
Kimi K2.5&R&C&.457&.513&.424&.009&.011&.051&.000&.000&.000\\
\rowcolor{CVTPaleGreen}
Qwen 3.5 OS&R&C&.389&.485&.423&.007&.057&.096&.000&.010&.020\\
\rowcolor{CVTPaleGreen}
Qwen 3.5 Plus&R&C&.416&.474&.415&.006&.050&.084&.000&.010&.010\\
\bottomrule
\end{tabular}}
\captionof{table}{\textbf{Complete ambiguity-aware results.} S/R = CVT-Synthetic/CVT-Real;
I/C = CVT-Isolated/CVT-Competing; Img/BB/SG = Image/Text-BBox/Scene Graph.
GPT-5.2 uses 10 competing scenes; all other models use 20.}
\label{tab:supp_all_results}
\end{figure*}

\begin{figure*}[t]
\centering
\includegraphics[width=\textwidth]{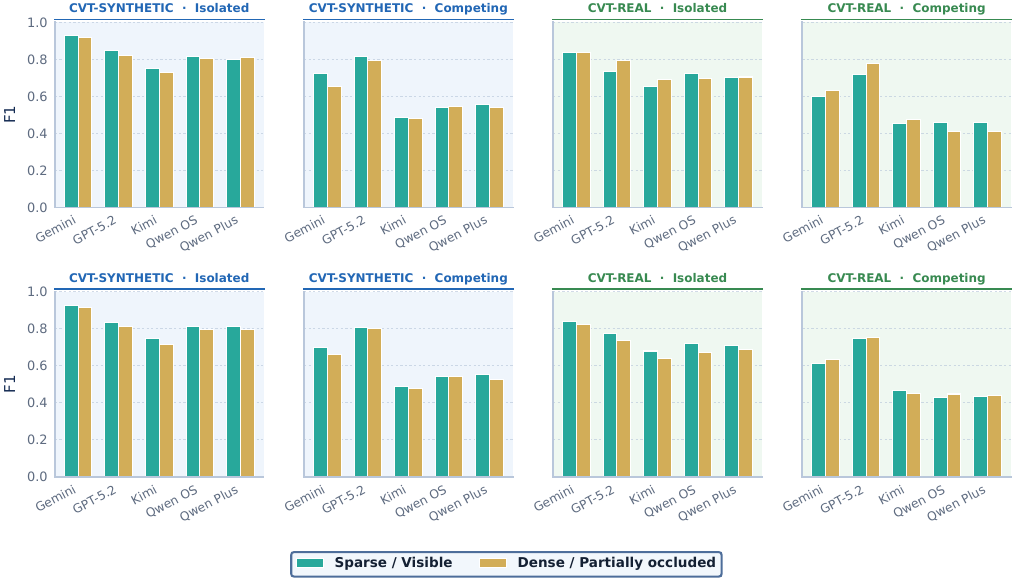}
\caption{\textbf{Difficulty slices.} Top row compares sparse and dense scenes; bottom row
compares questions involving only visible objects with those involving at least one partially
occluded object. Values average three representations within each model. Density and visibility
produce modest, nonuniform effects relative to the large regime and viewpoint effects.}
\label{fig:supp_difficulty}
\end{figure*}
\begin{figure*}[t]
\centering
\includegraphics[width=\textwidth]{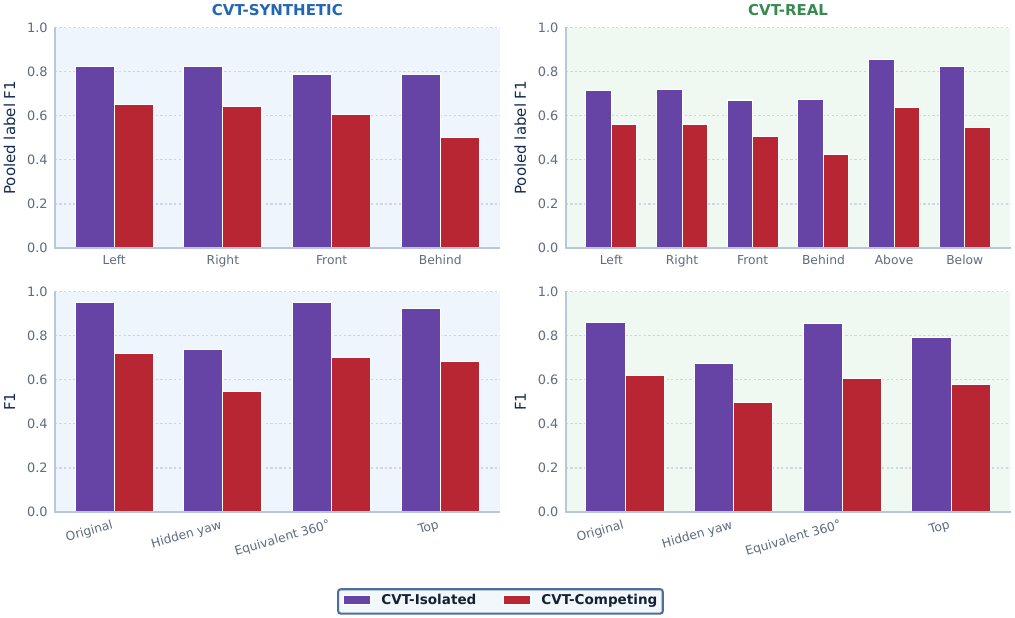}
\caption{\textbf{Relation-family and viewpoint diagnostics.} Relation F1 (top) and view-family
F1 (bottom), averaged over models and representations. Quarter turns remain difficult in both
domains and regimes. CVT-Real vertical predicates are more accurate and persistent because they
remain invariant under yaw, while horizontal predicates require viewpoint transformation.}
\label{fig:supp_relation_view}
\end{figure*}
The fixed-tracker-only angular curves differ from full curves by at most .81/.66 F1 points on
CVT-Synthetic and .48/1.21 on CVT-Real under CVT-Isolated/CVT-Competing. Hence, viewpoint
trends are not produced by changes in queried pairs. The counterfactual gap is .166/.139 on
CVT-Synthetic and .151/.099 on CVT-Real. Cycle agreement is 99.1\%/85.8\% and
97.9\%/84.9\%, respectively. The fraction for which both \(0^\circ\) and \(360^\circ\) are
correct falls from 87.5\% to 45.6\% on CVT-Synthetic and 62.0\% to 28.9\% on CVT-Real.

\subsection{Load-matched analysis excluding GPT-5.2}

GPT-5.2's C10 prompts are not load-matched to the C20 prompts used for the other models. We
therefore repeat the aggregate comparison on the four-model C20 cohort only. This is not a new
ranking table: it tests whether mixing GPT-5.2's lower-load results weakens or creates the reported
regime effect. GPT-5.2's repeated inability to return a complete C20 response remains an observed
completion failure in its own right, but is kept separate from relational correctness and coherence.

\begin{center}
\begin{minipage}{\columnwidth}
\centering
\small
\setlength{\tabcolsep}{4pt}
\begin{tabular}{llrr}
\toprule
\textbf{Domain} & \textbf{Metric} & \textbf{All-model drop} & \textbf{C20-only drop}\\
\midrule
Synthetic & F1 & .208 & .253\\
Synthetic & NSC & .266 & .299\\
Synthetic & Realizability & .349 & .387\\
Real & F1 & .198 & .244\\
Real & NSC & .162 & .182\\
Real & Realizability & .245 & .265\\
\bottomrule
\end{tabular}
\captionof{table}{\textbf{Load-matched sensitivity.} C20-only excludes GPT-5.2 and averages the
same four C20 models across regimes. Every decline becomes larger, so the mixed-load GPT-5.2
results attenuate rather than create the headline effect.}
\label{tab:supp_c20_only}
\end{minipage}
\end{center}

GPT-5.2 remains informative as a lower-load within-model replication. From CVT-Isolated to its
C10 CVT-Competing condition, mode-averaged F1 changes by only .026 on CVT-Synthetic and .015
on CVT-Real, while NSC drops by .132/.084 and realizability drops by .200/.163. Thus, even where
GPT-5.2 largely preserves local accuracy, its predictions become less persistent and less globally
coherent. We do not use these C10 values as evidence of C20 equivalence.

\begin{figure*}[t]
\centering
\includegraphics[width=\textwidth]{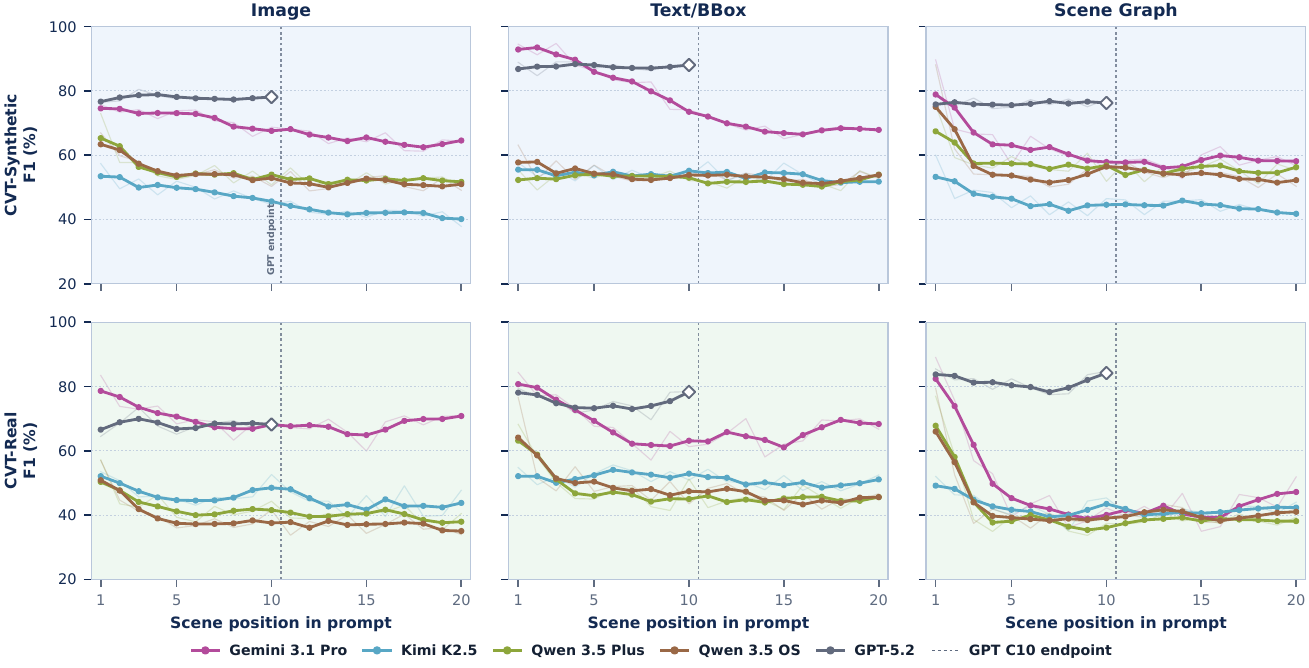}
\caption{\textbf{Position-wise CVT-Competing F1.} Faint paths show raw position scores and bold
paths show three-position rolling means. GPT-5.2 uses C10 and ends at the diamond after position
10; all other models use C20 and continue through position 20. The shared horizontal axis supports
within-range inspection but is not a load-matched cross-model comparison.}
\label{fig:supp_positionwise}
\end{figure*}

Figure~\ref{fig:supp_positionwise} exposes every within-prompt position rather than comparing only
aggregate C10 and C20 scores. GPT-5.2 remains broadly stable across the ten positions it encounters
in all six domain--representation panels, so its aggregate does not hide a sharp late-position
collapse within C10. Several C20 trajectories decline over their first positions and then flatten or
continue degrading. This analysis is descriptive, not a substitute for evaluating every model at
the same total scene load: shared position indices do not equalize total context or scene grouping.
It therefore clarifies GPT-5.2's observed C10 behavior without treating it as a C20 result.

\subsection{Horizontal and vertical tracking in CVT-Real}

\begin{center}
\begin{minipage}{\columnwidth}
\centering
\small
\setlength{\tabcolsep}{4pt}
\resizebox{\columnwidth}{!}{%
\begin{tabular}{lrr}
\toprule
\textbf{Tracker component} & \textbf{Isolated F1 / NSC} & \textbf{Competing F1 / NSC}\\
\midrule
2D trackers & .711 / .284 & .538 / .101\\
3D horizontal component & .688 / .298 & .478 / .132\\
3D vertical component & .851 / .769 & .601 / .834\\
\bottomrule
\end{tabular}}
\captionof{table}{\textbf{CVT-Real tracking decomposition.} Persistence loss concentrates in
viewpoint-remapped horizontal relations; above/below remain invariant under yaw.}
\label{tab:supp_real_3d}
\end{minipage}
\end{center}

\subsection{Can hidden viewpoints be inferred from one source view?}

Human evaluators receive exactly the source evidence given to models and never observe a target
view. After excluding observed \(0^\circ\) questions, mean ambiguity-aware F1 remains 83.7\% on
CVT-Synthetic and 81.6\% on CVT-Real. Thus, most hidden-view relations remain recoverable at the
benchmark's relational granularity even with natural depth, occlusion, and irregular object extent.
The small 2.1-point human domain difference shows that natural-scene construction does not cause a
broad loss of answerability. Together with the small strict-to-ambiguity-aware differences reported
above, this indicates that CVT-Real's main difficulty is counterfactual spatial inference rather than
pervasive target underdetermination.

\subsection{Separating source acquisition from transformation}

We next remove source-view correctness from the evaluation. First, a symbolic oracle that uses
the ground-truth source state and the benchmark transformation/query codec obtains 100\% F1 at
all ten viewpoints in both domains. This validates tag correspondence, query indexing, camera
conventions, and answer transformation. We then treat each model's own \(0^\circ\) prediction as
its believed canonical state, transform it symbolically, and compare the result with that model's
actual cardinal-view predictions. Consequently, a coherent but incorrect source interpretation
receives full credit.

\begin{center}
\begin{minipage}{\columnwidth}
\centering
\scriptsize
\setlength{\tabcolsep}{3pt}
\resizebox{\columnwidth}{!}{%
\begin{tabular}{llrrrrr}
\toprule
\textbf{Domain} & \textbf{Regime} & \textbf{Self \(90^\circ\)} & \textbf{Self \(180^\circ\)} &
\textbf{Self \(270^\circ\)} & \textbf{Self \(360^\circ\)} & \textbf{Perfect C4}\\
\midrule
Synthetic & Isolated  & 64.4 & 93.6 & 65.1 & 99.3 & 60.4\\
Synthetic & Competing & 29.6 & 49.7 & 27.9 & 87.2 & 18.5\\
Real      & Isolated  & 59.0 & 87.4 & 59.5 & 98.5 & 53.2\\
Real      & Competing & 25.6 & 49.0 & 26.9 & 85.6 & 17.8\\
\bottomrule
\end{tabular}}
\captionof{table}{\textbf{Model-relative transport (\%).} Each model's \(0^\circ\) answer is its
own anchor. Perfect C4 requires one coherent orbit across \(0^\circ,90^\circ,180^\circ,
270^\circ,360^\circ\), independent of benchmark correctness.}
\label{tab:supp_model_relative}
\end{minipage}
\end{center}

As a second, model-independent test, we retain a fixed pair only when every cardinal ground-truth
answer exactly equals the symbolic yaw transform of its \(0^\circ\) horizontal relation. This
keeps all 297 Synthetic trajectories and 426/494 (86.2\%) Real trajectories. Model predictions,
learned decoders, object categories, and human answers never affect selection. Diagonals and Top
are excluded from this audit because qualitative \(0^\circ\) labels need not uniquely determine a
diagonal projection.

\begin{center}
\begin{minipage}{\columnwidth}
\centering
\scriptsize
\setlength{\tabcolsep}{3pt}
\resizebox{\columnwidth}{!}{%
\begin{tabular}{llrrrr}
\toprule
\textbf{Domain} & \textbf{Regime} & \textbf{F1} & \textbf{Exact} &
\textbf{Perfect C4} & \textbf{Realizable}\\
\midrule
Synthetic & Isolated  & .836 & .750 & .539 & .577\\
Synthetic & Competing & .620 & .400 & .155 & .164\\
Real      & Isolated  & .769 & .520 & .347 & .479\\
Real      & Competing & .550 & .256 & .110 & .149\\
\bottomrule
\end{tabular}}
\captionof{table}{\textbf{Algebraically recoverable cardinal subset.} Values average all saved
model--representation cells. Real Isolated-to-Competing drops remain 21.9 F1, 23.7 perfect-C4,
and 33.0 realizability points after conservative filtering.}
\label{tab:supp_recoverable_subset}
\end{minipage}
\end{center}

These complementary gates separate acquisition from transformation: perfect oracle performance
rules out codec errors; model-relative transport removes disagreement with source-view ground truth;
and GT-only filtering removes non-closed cardinal trajectories. All retain the same collapse under
competing spatial contexts. Source-view difficulty therefore cannot explain the persistence and
coherence results, although individual Real questions may still vary in perceptual difficulty.

\subsection{Why quarter turns remain difficult}

We audit the recurring \(90^\circ/270^\circ\) dip with fixed pairs and frozen predictions.
Every yaw contains the same tracked scene--pair families. Quarter-turn questions occur .95
questions earlier than other views on CVT-Synthetic and .08 earlier on CVT-Real; angle explains
only .14\% and .24\% of prompt-position variance. Equivalent short-\(90^\circ\) and
long-\(270^\circ\) wordings differ by at most 1.7 exact points. Pair composition, placement,
and wording therefore do not explain the pattern.

\begin{center}
\begin{minipage}{\columnwidth}
\centering
\scriptsize
\setlength{\tabcolsep}{2pt}
\resizebox{\columnwidth}{!}{%
\begin{tabular}{llrrrr}
\toprule
\textbf{Domain} & \textbf{Regime} & \textbf{Diagonal exact / F1} &
\textbf{Quarter exact / F1} & \textbf{\(180^\circ\) exact} &
\textbf{\(180^\circ-\)quarter exact} \\
\midrule
Synthetic & Isolated & .455 / .720 & .573 / .663 & .818 & .245 \\
Synthetic & Competing & .265 / .537 & .295 / .515 & .395 & .100 \\
Real & Isolated & .274 / .592 & .375 / .582 & .525 & .150 \\
Real & Competing & .164 / .438 & .203 / .431 & .270 & .067 \\
\bottomrule
\end{tabular}}
\captionof{table}{\textbf{Exact/F1 decomposition by transformation.} Diagonal errors retain more
partial-axis credit, which can make their F1 appear higher despite lower exact-set accuracy.
Quarter turns remain substantially harder than \(180^\circ\) under exact scoring.}
\label{tab:supp_quarter_exact}
\end{minipage}
\end{center}
The algebra explains the distinction: \(R_{180}(x,y)=(-x,-y)\) independently reverses two
axes, whereas \(R_{90}(x,y)=(y,-x)\) and \(R_{270}(x,y)=(-y,x)\) exchange semantic axes and
bind sign to rotation direction. Within the same diagonal angles and after requiring a correct
source answer, relations whose ground truth requires the simpler \(0^\circ/180^\circ\) endpoint
are 4.9/11.4 points easier than cross-axis endpoints on CVT-Synthetic and 5.6/11.1 points
easier on CVT-Real under Isolated/Competing; 14/15, 13/15, 14/15, and 15/15 cells show the
same direction. This is behavioral evidence for axis-rebinding difficulty, not a claim about
a unique internal algorithm.

\section{Robustness and Statistical Inference}

\begin{figure*}[t]
\centering
\includegraphics[width=\textwidth]{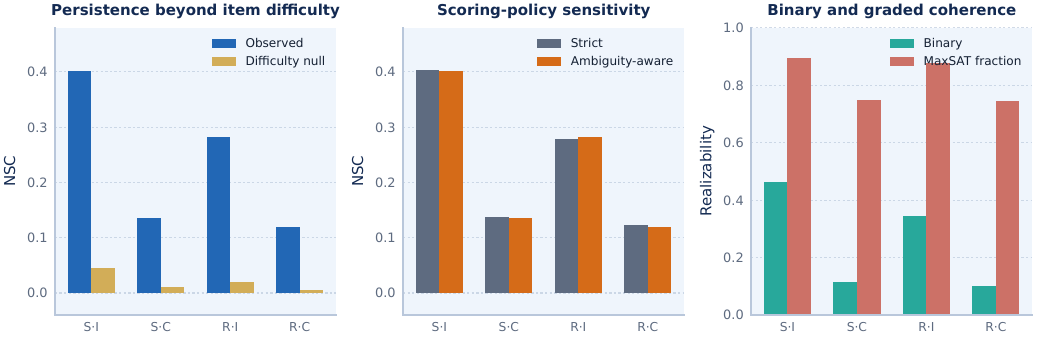}
\caption{\textbf{Metric robustness.} (a) Observed NSC substantially exceeds an item-difficulty
null. (b) Strict and ambiguity-aware F1 are nearly identical. (c) Graded MaxSAT remains high but
drops under competition, complementing strict binary realizability. Bars average model and
representation cells; error bars show across-cell standard error where applicable.}
\label{fig:supp_robustness}
\end{figure*}
\subsection{Persistence beyond item difficulty}

We independently permute trajectory identity within each encounter step, preserving every
step's marginal difficulty while destroying across-step persistence. Five hundred permutations
are performed per model--representation cell, followed by Benjamini--Hochberg correction across
60 cells.

\begin{center}
\begin{minipage}{\columnwidth}
\centering
\small
\setlength{\tabcolsep}{4pt}
\begin{tabular}{llrr}
\toprule
\textbf{Domain} & \textbf{Regime} & \textbf{Observed NSC} & \textbf{Null NSC}\\
\midrule
Synthetic & Isolated & .401 & .045\\
Synthetic & Competing & .136 & .011\\
Real & Isolated & .282 & .020\\
Real & Competing & .120 & .005\\
\bottomrule
\end{tabular}
\captionof{table}{\textbf{Item-difficulty null.} Fifty-two of 60 cells remain significant at
BH-adjusted \(q<.05\); observed persistence is not reducible to easy items recurring together.}
\label{tab:supp_nsc_null}
\end{minipage}
\end{center}
\subsection{Prompt-clustered paired inference}

Questions within one API response are dependent. We therefore resample complete competing
prompt batches and matched isolated scene blocks using 5,000 paired draws. All aggregate
intervals exclude zero.

\begin{center}
\begin{minipage}{\columnwidth}
\centering
\scriptsize
\setlength{\tabcolsep}{2pt}
\resizebox{\columnwidth}{!}{%
\begin{tabular}{llrrrrr}
\toprule
\textbf{Domain} & \textbf{Metric} & \textbf{Isolated} & \textbf{Competing} &
\textbf{Drop} & \textbf{95\% CI} & \textbf{Cell CIs \(>0\)}\\
\midrule
Synthetic & F1 & .821 & .613 & .208 & [.192,.224] & 12/15\\
Synthetic & NSC & .401 & .136 & .266 & [.250,.299] & 14/15\\
Synthetic & Binary realizability & .463 & .114 & .349 & [.313,.387] & 12/15\\
Synthetic & MaxSAT & .894 & .747 & .146 & [.132,.161] & 12/15\\
Real & F1 & .737 & .539 & .198 & [.186,.211] & 12/15\\
Real & NSC & .282 & .120 & .162 & [.134,.196] & 10/15\\
Real & Binary realizability & .345 & .101 & .245 & [.206,.282] & 10/15\\
Real & MaxSAT & .878 & .745 & .133 & [.118,.147] & 12/15\\
\bottomrule
\end{tabular}}
\captionof{table}{\textbf{Prompt-clustered uncertainty.} GPT-5.2 contributes ten 10-scene prompt
clusters; other competing cells contribute five 20-scene clusters. Aggregate effects are robust,
while cell-level intervals are reported without treating questions as independent.}
\label{tab:supp_cluster_ci}
\end{minipage}
\end{center}
\subsection{Coherence beyond local error rate}

We exactly match isolated and competing outputs by model, representation, local error count,
and, in a stricter comparison, asserted-constraint count. Realizability still decreases by .163
on CVT-Synthetic (95\% CI [.080,.197]) and .189 on CVT-Real ([.047,.236]). Matching only
error count yields drops of .203 ([.139,.237]) and .164 ([.127,.194]). Constraint count has
small correlation with binary realizability: \(-.043/-.039\) on CVT-Synthetic and
\(-.018/+.117\) on CVT-Real under Isolated/Competing. The coherence collapse therefore is not
an artifact of more wrong or more verbose predictions.

Binary feasibility is intentionally strict, so we also report MaxSAT. Mean satisfiable fraction
falls from .894 to .747 on CVT-Synthetic and .878 to .745 on CVT-Real. Binary realizability
equals the event MaxSAT \(=1\) for every one of 6,000 scene outputs, and every optimization
instance terminates successfully.

\subsection{Encounter-order and prompt-partition sensitivity}

Post-hoc random reordering of frozen answers changes aggregate NSC by only .008/.019 on
CVT-Synthetic and \(-.001/-.001\) on CVT-Real under Isolated/Competing; mean absolute cell
changes are .013/.020 and .009/.009. Thus, the saved shuffled encounter order is not responsible
for the domain or regime conclusions.

Across existing prompt partitions, 71/90 synthetic and 73/90 real prompt blocks show positive
F1 drops. Twelve of 15 cells in each domain are positive in every block; leave-one-block-out
ranges preserve direction in 28/30 cells. With only five 20-scene blocks, exact cell-level sign
tests are low-powered, motivating the aggregate prompt-clustered intervals above.

\subsection{Counterfactual-only and output-integrity audits}

Excluding \(0^\circ\), \(360^\circ\), and Top, mean hidden-yaw F1 remains .739/.550 on
CVT-Synthetic and .674/.499 on CVT-Real under CVT-Isolated/CVT-Competing. Headline trends
therefore are not driven by observed or equivalent views.

Across 1,590 CVT-Real jobs, 1,713 attempts were recorded. Ninety jobs required more than one
attempt; 108 attempts ended in retryable provider/transport errors and 17 attempt records were
parse-invalid or incomplete. Only complete, validated outputs are retained. Canonical evaluation
contains no missing scenes, missing answers, key collisions, truncations, provider mismatches,
or payload mismatches. Retries affect cost and latency, not which incomplete outputs are scored.

\FloatBarrier
\section{Matched Context Controls}

The controls form a staged intervention ladder rather than two interchangeable filler conditions.
Each prompt retains exactly one target spatial scene at position 0, 9, or 19; these positions are
evaluated in separate prompts, never as three targets in one prompt. Target identity, spatial
questions, and placement remain fixed across the corresponding conditions.

\noindent\textbf{Irrelevant filler.}
The other 19 scene blocks are replaced by repeated semantically irrelevant text matched to the
removed text length. These slots contain no images, questions, or required reasoning and can be
ignored by the model. This condition asks whether the same target degrades merely because it occurs
later within an equally long text context, without competing scenes or intervening reasoning.

\noindent\textbf{Attribute filler.}
The other 19 slots retain their actual Image or Text/BBox scene data and their original episode
ordering, but counterfactual spatial questions are replaced by low-complexity color, shape,
material, object, and count retrieval questions whose answers are directly available in each scene
dictionary. The generator exactly matches each replaced scene's question count and one/two-answer
cardinality distribution; aggregate prompt character burden remains within 5\% of CVT-Competing.
Thus, this control preserves scene switching, visual and representational load, question/output
structure, target position, and the need to access every scene. Low reasoning complexity is
deliberate: it tests whether models retain the target while performing genuine but directly
grounded work on surrounding scenes. Matching counterfactual transformation difficulty would
instead recreate the CVT-Competing treatment.

\noindent\textbf{CVT-Competing.}
All 20 scene episodes require counterfactual spatial reasoning. The three stages therefore separate
an ignorable long context, a matched multi-scene low-reasoning workload, and competing spatial
transformations. We evaluate 15 matched target prompts per model and mode.

Paired target-prompt intervals for irrelevant filler minus competing F1 are [.120,.256] for
Gemini and [.114,.246] for Qwen; NSC intervals are [.100,.711] and [\(-.004\),.457]. Attribute
filler minus competing is uncertain for Gemini (F1 [\(-.149\),.094], NSC [\(-.169\),.467]) but
positive for Qwen (F1 [.038,.170], NSC [.054,.379]).

\begin{table}[t]
\centering
\scriptsize
\setlength{\tabcolsep}{2pt}
\begin{tabular}{llrrrr}
\toprule
\textbf{Model} & \textbf{Metric} & \textbf{Iso.} & \textbf{Comp.} &
\textbf{Irr.} & \textbf{Attr.}\\
\midrule
Gemini & F1 & .955 & .695 & .883 & .677\\
Gemini & NSC & .787 & .293 & .664 & .408\\
Qwen OS & F1 & .700 & .552 & .735 & .657\\
Qwen OS & NSC & .180 & \(-.013\) & .200 & .179\\
\bottomrule
\end{tabular}
\caption{\textbf{Matched Image-mode target results.} Values pool the same 15 target scenes;
Irr./Attr. denote irrelevant/attribute filler.}
\label{tab:supp_fillers}
\end{table}

\begin{table}[t]
\centering
\scriptsize
\resizebox{\columnwidth}{!}{%
\begin{tabular}{lrrrrrrrrrr}
\toprule
\textbf{Model / mode} & \textbf{0}&\textbf{1}&\textbf{2}&\textbf{3}&\textbf{4}&\textbf{5}&\textbf{6}&\textbf{7}&\textbf{8}&\textbf{9}\\
\midrule
Gemini Image&90.0&91.9&92.3&92.3&94.8&92.3&94.8&90.4&94.9&99.0\\
Gemini Text/BBox&90.7&94.0&93.9&93.1&93.2&92.2&92.3&92.9&93.1&99.8\\
Qwen OS Image&95.9&95.5&95.8&94.3&93.8&96.4&94.8&95.3&92.2&98.7\\
Qwen OS Text/BBox&76.6&73.0&69.5&70.0&70.9&73.7&73.1&72.6&71.5&68.3\\
\bottomrule
\end{tabular}}
\caption{\textbf{Attribute retrieval by batch position (\%), positions 0--9.}}
\label{tab:supp_retrieval_position_a}
\end{table}

\begin{table}[t]
\centering
\scriptsize
\resizebox{\columnwidth}{!}{%
\begin{tabular}{lrrrrrrrrrrr}
\toprule
\textbf{Model / mode} & \textbf{10}&\textbf{11}&\textbf{12}&\textbf{13}&\textbf{14}&\textbf{15}&\textbf{16}&\textbf{17}&\textbf{18}&\textbf{19}&\textbf{All}\\
\midrule
Gemini Image&92.1&92.1&93.5&94.7&92.4&93.3&91.7&91.4&94.2&92.3&93.0\\
Gemini Text/BBox&95.3&96.0&92.5&91.9&95.0&92.9&93.1&92.5&95.2&90.0&93.5\\
Qwen OS Image&96.3&96.8&96.2&95.8&96.1&96.9&95.7&94.8&94.7&94.5&95.5\\
Qwen OS Text/BBox&70.0&70.0&73.2&75.4&72.2&71.9&73.0&70.3&69.3&69.8&71.7\\
\bottomrule
\end{tabular}}
\caption{\textbf{Attribute retrieval, positions 10--19.} Accuracy remains stable rather than
decaying with target position; Qwen Text/BBox is lower overall but likewise does not collapse.}
\label{tab:supp_retrieval_position}
\end{table}

\begin{figure*}[t]
\centering
\includegraphics[width=\textwidth]{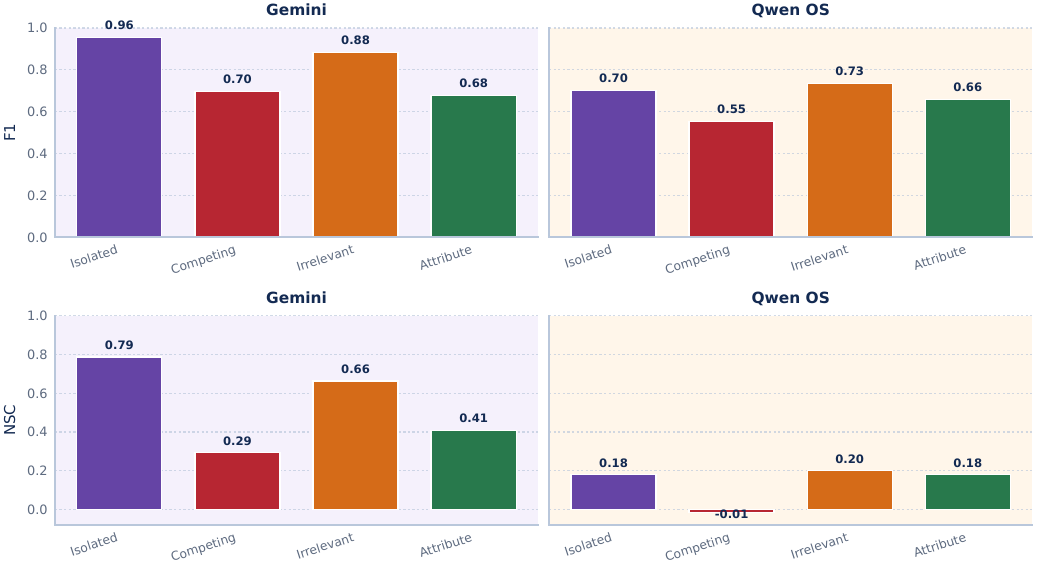}
\caption{\textbf{Matched context controls.} Target-scene F1 and NSC for Gemini 3.1 Pro and
Qwen 3.5 OS in Image mode. Irrelevant filler recovers much of the loss despite comparable
prompt position and length. Attribute filler preserves scene access but produces model-dependent
recovery, separating generic context load from competing spatial content.}
\label{fig:supp_fillers}
\end{figure*}
\section{Qualitative Examples}

\begin{figure*}[t]
\centering
\includegraphics[width=0.92\textwidth]{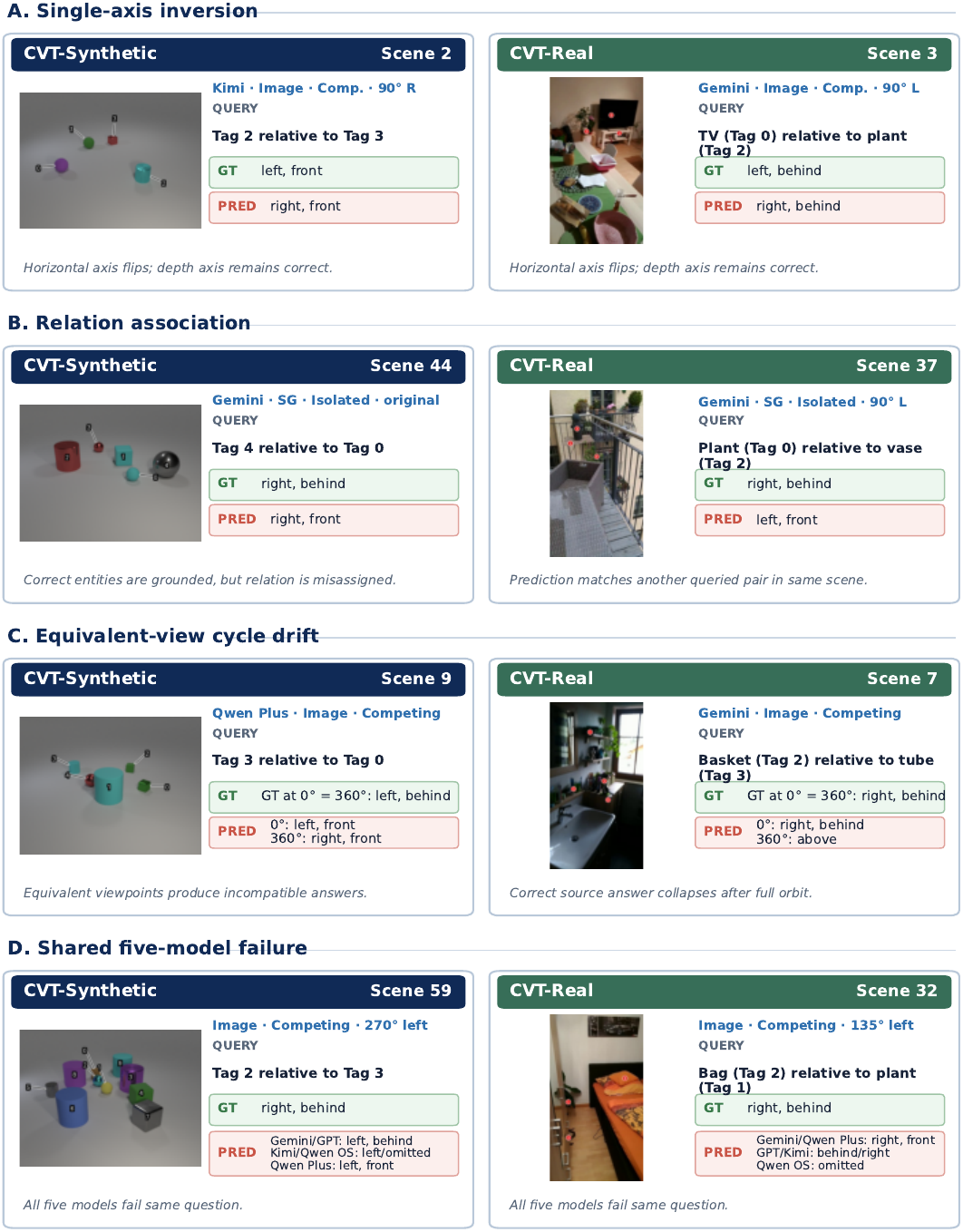}
\caption{\textbf{Matched qualitative failures across domains.} Every panel shows the tagged
source image actually observed by the model; no counterfactual target view is rendered. Rows pair
CVT-Synthetic and CVT-Real examples of single-axis inversion, relation association, equivalent-view
cycle drift, and questions missed by all five evaluated models. Predictions come from the frozen
benchmark outputs and use ambiguity-aware scoring.}
\label{fig:supp_qual_images}
\end{figure*}

\noindent\textbf{F.1\quad Human validation protocol.}
Evaluators receive the same zoomable tagged source view, tag--object dictionary, question
wording, and question order used for Image-mode model evaluation. They never observe a target
view, answer every benchmark question for the selected scene, and receive no correctness
feedback. We retain the full scene question set rather than removing difficult views or pairs.
The selected scenes span sparse and dense layouts and controlled and natural geometry.

Targeted validation covers three scenes of diverse difficulty per domain, three annotators,
and roughly 180 questions per evaluator in CVT-Isolated. Mean F1 is 84.9\% on CVT-Synthetic
and 81.8\% on CVT-Real, versus matched five-model means of 79.9\% and 70.7\%. These studies
validate task solvability and ground-truth quality; they are not human ceilings.

\noindent\textbf{F.2\quad Failure reading.}
Figure~\ref{fig:supp_qual_images} holds the observed source image and directed query explicit while
showing four recurring behaviors. Single-axis inversions preserve one component but reverse the
other; association errors attach a valid relation to the wrong directed pair; cycle drift gives
incompatible answers at equivalent views; and shared hard cases defeat all five models. Their
presence in both domains indicates failures beyond one renderer, object vocabulary, or scene type.

\FloatBarrier
\section{Models, Prompts, and Reproducibility}

\subsection{Models and system profile}

\begin{table}[t]
\centering
\scriptsize
\setlength{\tabcolsep}{2pt}
\renewcommand{\arraystretch}{1.08}
\begin{tabularx}{\columnwidth}{@{}lXl@{}}
\toprule
\textbf{Model} & \textbf{Identifier / official provider} & \textbf{Use}\\
\midrule
Gemini 3.1 Pro & google/gemini-3.1-pro-preview; Google AI Studio & I/C + controls\\
GPT-5.2 & openai/gpt-5.2; OpenAI & I + C10\\
Kimi K2.5 & moonshotai/kimi-k2.5; Moonshot & I + C20\\
Qwen 3.5 OS & qwen/qwen3.5-397b-a17b; Alibaba & I/C + controls\\
Qwen 3.5 Plus & qwen/qwen3.5-plus-02-15; Alibaba & I + C20\\
\bottomrule
\end{tabularx}
\caption{\textbf{Model and inference configuration.} I/C = CVT-Isolated/CVT-Competing.
Models use official provider-recommended high-capability configurations. Real and filler calls
pin official providers and disable fallback.}
\label{tab:supp_models}
\end{table}

\begin{table}[t]
\centering
\scriptsize
\setlength{\tabcolsep}{1.5pt}
\resizebox{\columnwidth}{!}{%
\begin{tabular}{lrrrr}
\toprule
\textbf{Model} & \textbf{Context window} & \textbf{Output cap} & \textbf{Peak input tokens} & \textbf{Context used}\\
\midrule
Gemini 3.1 Pro & 1,000,000 & 65,536 & 96,507 & 9.7\%\\
GPT-5.2 & 400,000 & 128,000 & 41,205 & 10.3\%\\
Kimi K2.5 & 262,144 & 65,536 & 82,991 & 31.7\%\\
Qwen 3.5 OS & 262,144 & 32,768 & 86,772 & 33.1\%\\
Qwen 3.5 Plus & 262,144 & 32,768 & 86,772 & 33.1\%\\
\bottomrule
\end{tabular}}
\caption{\textbf{Context utilization.} Peak inputs occupy 9.7--33.1\% of advertised context
windows. GPT-5.2 uses 10-scene competing prompts; all other models use 20.}
\label{tab:supp_context}
\end{table}
The full CVT-Real run uses global concurrency 64, per-model concurrency 16, and per-provider
concurrency 24. Each job permits five attempts with exponential backoff, cooldown, start jitter,
response validation, incomplete-output retry, output locks, atomic writes, and resumable request
hashes. Every retained generation records model/provider identifiers, request hash, response
metadata, attempts, and token usage. Table~\ref{tab:supp_context} shows that peak inputs occupy
only 9.7--33.1\% of advertised context windows.

\subsection{Analysis environment and artifact inventory}

Canonical analysis uses Python 3.12.4, NumPy 1.26.4, SciPy 1.13.1, pandas 2.2.2,
scikit-learn 1.4.2, Matplotlib 3.11.0, aiohttp 3.14.1, and Pillow 12.2.0.
Spatial feasibility uses \texttt{scipy.optimize.linprog}; graded realizability uses
\texttt{scipy.optimize.milp}. Benchmark construction and all audits use frozen random seeds.

Release artifacts include manifests and tagged inputs, prompts and provider settings, validated
predictions, ambiguity sidecars, the evaluator, realizability and MaxSAT solvers, audit and plot
scripts, and machine-readable values. Representative files accompany submission; the complete
benchmark and code will follow acceptance.

\subsection{Prompt templates}

All models receive the same task content and output schema. The canonical prompts are given
below. Provider-specific adaptations only shorten wording or explicitly require every item to
be answered; they do not alter scene evidence, questions, relation options, or expected outputs.

\begin{promptblock}{Image master prompt}
\begin{lstlisting}[style=cvtprompt]
You will be given a series of images with questions about spatial relationships between tagged objects. Answer all questions based on careful spatial reasoning. Format your answers as a single JSON: {'<image_name>': {'q_<id>': [relationships], ...}, ...}.
\end{lstlisting}
\end{promptblock}

\begin{promptblock}{Text/BBox master prompt}
\begin{lstlisting}[style=cvtprompt]
You will be given a series of scene descriptions with questions about spatial relationships between tagged objects. Answer all questions based on careful spatial reasoning. Format your answers as a single JSON: {'<image_name>': {'q_<id>': [relationships], ...}, ...}.
\end{lstlisting}
\end{promptblock}

\begin{promptblock}{Scene Graph master prompt}
\begin{lstlisting}[style=cvtprompt]
You will be given a series of scenes. Each scene contains:
  1. A Scene Object Dictionary: maps Tag IDs to object descriptions and bounding boxes [x, y, w, h]. No image is provided.
  2. A Scene Graph for the original view (0deg).
     Format: 'Tag X | dir:[A, B]' means Tags A and B are in direction 'dir' relative to Tag X.
     e.g. 'Tag 0 | left:[1, 3]' means Tag 1 and Tag 3 are to the LEFT of Tag 0.
     Only directions with targets are listed; unlisted directions have no targets.

Answer all spatial relationship questions using the scene graph and object dictionary. For rotation questions, mentally rotate the scene from the original configuration.
Format your answers as a single JSON: {'<image_name>': {'q_<id>': [relationships], ...}, ...}.
\end{lstlisting}
\end{promptblock}

\subsection{Scene blocks and question templates}

After the master prompt, each scene contributes its complete representation-specific context
followed immediately by its questions. The same scene JSON produces all three modes.
To keep instruction wording uniform across scenes and regimes, executed scene blocks use the
same nominal ``60 questions'' phrase even when sparse or vertically enriched scenes contain a
different count. Every actual question is explicitly and contiguously numbered, the output
schema requests all listed IDs, and retained-output validation confirms that models answered
the complete enumerated list; the nominal phrase is never used for parsing or scoring.

\begin{promptblock}{Image scene block}
\begin{lstlisting}[style=cvtprompt]
Questions for Image '<image_name>':
Refer to the numeric tags overlaid on the objects in the image.
Scene Object Dictionary: {"0": "<description>", ...}
Answer the following 60 questions based on the image and object dictionary:
\end{lstlisting}
\end{promptblock}

\begin{promptblock}{Text/BBox scene block}
\begin{lstlisting}[style=cvtprompt]
Scene '<image_name>':
Refer to the bounding boxes [x, y, w, h] provided in the dictionary.
Scene Object Dictionary: {"0": ("<description>", [x,y,w,h]), ...}
Answer the following 60 questions based on the bounding boxes and object dictionary:
\end{lstlisting}
\end{promptblock}

\begin{promptblock}{Scene Graph scene block}
\begin{lstlisting}[style=cvtprompt]
Scene Object Dictionary: {"0": ("<description>", [x,y,w,h]), ...}
Scene Graph (original view):
Tag 0 | left:[1,3]  front:[1,2]
...
\end{lstlisting}
\end{promptblock}

Question wording is shared across representations and regimes; only relation options change for
Top and CVT-Real vertical questions.

\begin{promptblock}{Exact question templates}
\begin{lstlisting}[style=cvtprompt]
[SOURCE VIEW]
In the original view, what is the spatial relationship of the <object> (Tag <B>) with respect to the <subject> (Tag <A>)? Choose from [left, right, front, behind]. Multiple relations possible.

[COUNTERFACTUAL ORBIT]
If the camera orbits <angle> degrees to the <left/right>, what is the new spatial relationship of the <object> (Tag <B>) with respect to the <subject> (Tag <A>)? Choose from [left, right, front, behind]. Multiple relations possible.

[TOP VIEW]
If the camera orbits to a top-down view, what is the new spatial relationship of the <object> (Tag <B>) with respect to the <subject> (Tag <A>)? Choose from [left, right, south, north]. Multiple relations possible.
\end{lstlisting}
\end{promptblock}

CVT-Real vertical-enriched templates append above/below to the option list and preserve the
same directed-pair semantics. Internal north/south answers use the frozen benchmark codec
north \(\leftrightarrow\) behind and south \(\leftrightarrow\) front only at the evaluator boundary.
\end{document}